\documentclass[aoas]{imsart}

\RequirePackage{amsthm,amsmath,amsfonts,amssymb}
\RequirePackage[authoryear]{natbib}
\RequirePackage{graphicx}
\usepackage[plain,noend, ruled]{algorithm2e}
\usepackage{comment}
\graphicspath{{./art/}}
\usepackage{enumitem}
\usepackage{multirow}
\usepackage{booktabs}
\usepackage{longtable}
\usepackage{diagbox}

\usepackage{xr-hyper}
\usepackage[colorlinks]{hyperref}
\hypersetup{
    colorlinks,
	linkcolor={magenta},
    filecolor={cyan},
	citecolor={blue},
}

\newcommand{\E}{\mathbb{E}}
\newcommand{\Mean}{{\mathbb{E}}}

\startlocaldefs
\theoremstyle{plain}
\newtheorem{axiom}{Axiom}

\newtheorem{theorem}{Theorem}
\newtheorem{lemma}[axiom]{Lemma}

\theoremstyle{remark}

\newtheorem{assumption}{Assumption}
\newtheorem{remark}{Remark}

\endlocaldefs

\begin{document}

\begin{frontmatter}
\title{A Two-armed Bandit Framework for A/B Testing}
\runtitle{A Two-armed Bandit Framework for A/B Testing}

\begin{aug}
\author[A]{\fnms{Jinjuan}~\snm{Wang} \ead[label=e1]{wangjinjuan@bit.edu.cn}},
\author[B]{\fnms{Qianglin}~\snm{Wen} \ead[label=e2]{qianglin@mail.ynu.edu.cn}},
\author[C]{\fnms{Yu}~\snm{Zhang} \ead[label=e3]{202412074@mail.sdu.edu.cn}},
\author[D]{\fnms{Xiaodong}~\snm{Yan}\ead[label=e4]{yanxiaodong@xjtu.edu.cn}}$^{\color{magenta}\ast}$
\and
\author[E]{\fnms{Chengchun}~\snm{Shi}\ead[label=e5]{c.shi7@lse.ac.uk}}\thanks{Co-corresponding authors.}

\address[A]{School of Mathematics and Statistics, Beijing Institute of Technology \printead[presep={ ,\ }]{e1}}

\address[B]{Yunnan Key Laboratory of Statistical Modeling and Data Analysis, Yunnan University \printead[presep={ ,\ }]{e2}} 

\address[C]{Zhongtai Securities Institute for Financial Studies, Shandong University, Jinan, China \printead[presep={ ,\ }]{e3}}

\address[D]{School of Mathematics and Statistics, Xi'an Jiaotong University \printead[presep={,\ }]{e4}}

\address[E]{Department of Statistical Science, London School of Economics and Political Science \printead[presep={,\ }]{e5}}

\end{aug}

\begin{abstract}
A/B testing is widely used in modern technology companies for policy evaluation and product deployment, with the goal of comparing the outcomes under a newly-developed policy against a standard control. Various causal inference and reinforcement learning methods developed in the literature are applicable to A/B testing. This paper introduces a two-armed bandit framework designed to improve the power of existing approaches. The proposed procedure consists of three main steps: (i) employing doubly robust estimation to generate pseudo-outcomes, (ii) utilizing a two-armed bandit framework to construct the test statistic, and (iii) applying a permutation-based method to compute the $p$-value. We demonstrate the efficacy of the proposed method through asymptotic theories, numerical experiments and real-world data from a ridesharing company, showing its superior performance in comparison to existing methods.
\end{abstract}


\end{frontmatter}

\section{Introduction}
This paper aims to develop effective A/B testing solutions across various industries, including internet companies such as Google, LinkedIn, X, and Meta, e-commerce platforms like Amazon, and two-sided marketplaces such as Airbnb. A/B testing has become the gold standard in these companies for policy evaluation and product deployment. For example, on traditional portal websites, it is common to assess a new version of a webpage (B) against the existing one (A) by randomly assigning visitors to either variant and then comparing an outcome of interest -- such as the click through rate -- to determine whether B outperforms A.

\begin{figure}[t]
    \centering
    \includegraphics[width=0.75\linewidth]{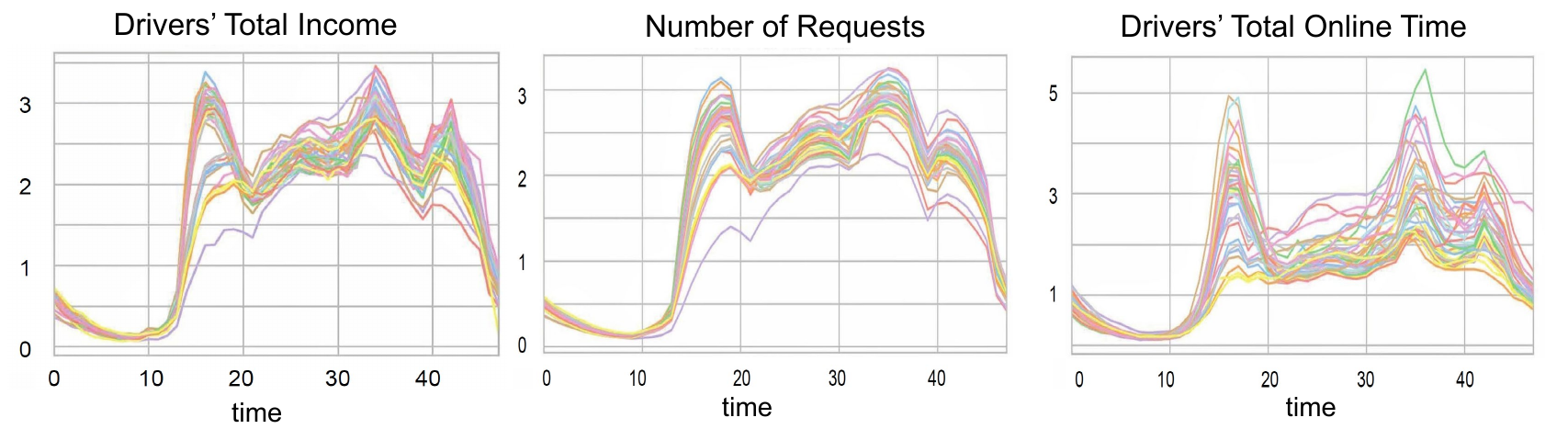}
    \includegraphics[width=0.75\linewidth]{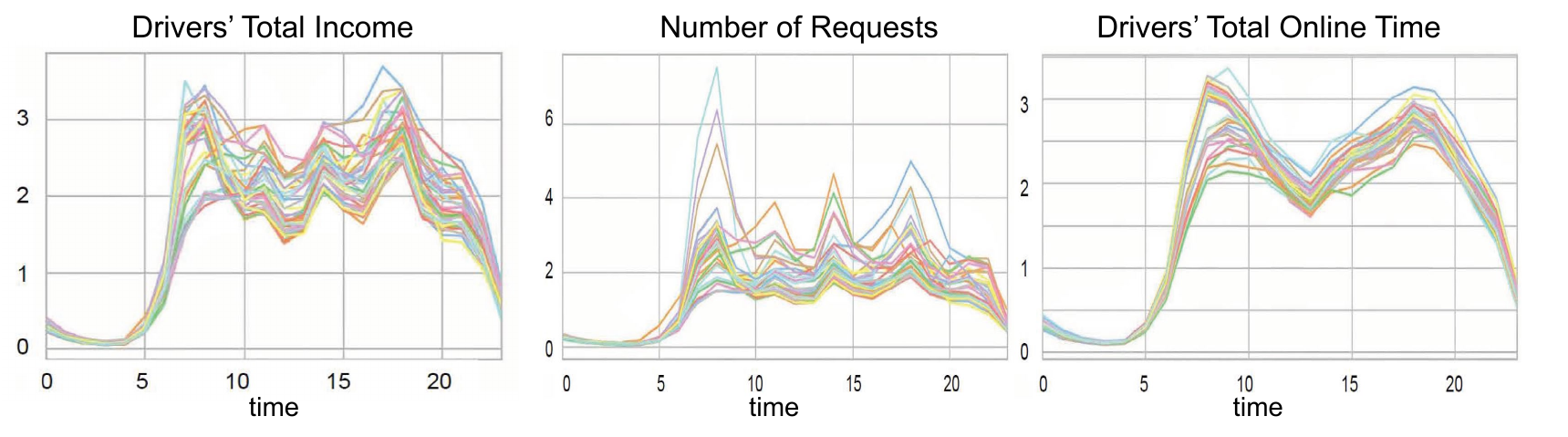}
    \caption{Drivers' total income, the numbers of call orders and drivers' total online time from two cities, taken from \citet{luo2022policy}. Each row presents data from one city. The values are scaled to preserve privacy.}\label{fig:drivertotalincome}
\end{figure}

A motivating application considered in this paper is the development of A/B testing solutions for large-scale fleet management in ride-sharing platforms, such as Uber and Lyft in the United States, and Didi Chuxing in China. The widespread adoption of smartphones and ride-sharing apps has enabled these companies to revolutionize and reshape urban transportation  \citep{alonso2017demand,hagiu2019status}. Ride-sharing platform is  a typical two-sided market that enables efficient interactions between passengers and drivers \citep{rysman2009economics}, as well as a complex spatio-temporal ecosystem \citep{wang2019ridesourcing}. Specifically, the demand and supply of this two-sided market can be measured by the numbers of call orders and the total drivers' online time in a city. These variables exhibit strong temporal patterns (see Figure \ref{fig:drivertotalincome} for a visualization), and interact with each other over time and location. 

Ride-sharing companies are particularly interested in evaluating the effects of two types of policies on various outcomes of interest, such as the answer rate (the percentage of call orders responded to by drivers), the completion rate (the percentage of call orders successfully completed), driver income, and gross merchandise value (GMV, the total transaction volume generated on the two-sided market through the ride-sharing platform). The first type of policy is the subsidy policy, which can be targeted at either drivers or passengers. For example, under a passenger-side subsidy policy, some passengers may receive coupons that offer discounts to call orders requested within a specified time frame. The goal of such policies is to encourage more call orders and to increase passenger engagement with the platform -- especially among new passengers. The second type of policy is the order dispatch policy, which focuses on assigning the most suitable available driver to each call order in the city. This is essentially a matching problem between supply and demand \citep[see e.g.,][]{xu2018large,tang2019deep,zhou2021graph}. Both types of policies affect platform outcomes (e.g., GMV) through their effects on the supply and demand. Specifically, passenger- and driver-side subsidy policies increase the GMV by stimulating more call orders and extending drivers' online time, respectively, while order dispatch policies affect the GMV by efficiently matching drivers to orders and optimizing their locations across the city. 

\begin{figure*}[t!]
\centering
\includegraphics[width=12.5cm]{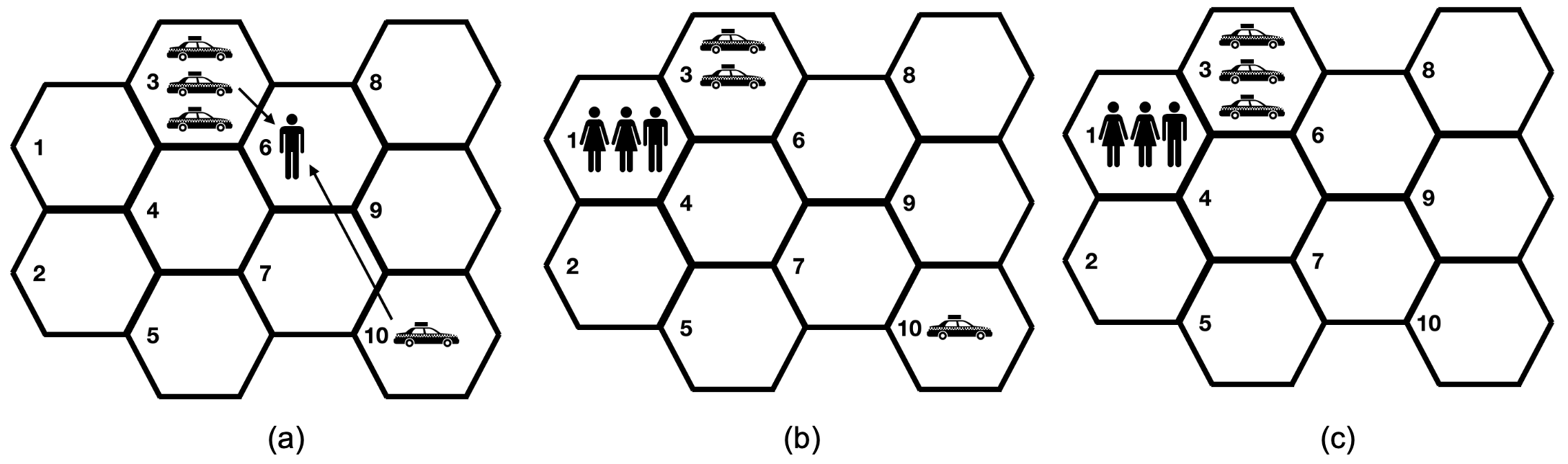}
\caption{Visualization of the carryover effect using a ride-sharing example, taken from \citet{li2024evaluating}. (a) The city is divided into ten regions, and a passenger in Region 6 orders a ride. Two actions are available: assigning a driver from Region 3 or from Region 10. These actions lead to different future outcomes, as illustrated in (b) and (c). (b) Assigning the driver from Region 3 may result in a future call order in Region 1 being canceled, since the driver in Region 10 is too far away. As a result, the passenger in Region 1 may cancel the order due to the long wait time. (c) Assigning the driver from Region 10 keeps all three drivers in Region 3 idle and available to fulfill the future three call orders from Region 1.}\label{fig:carryover}
\end{figure*}

A/B testing in modern technological industries poses several practical challenges. The first is the small sample size. Specifically, online experiments are typically constrained to a few weeks \citep{bojinov2023design}. For instance, when evaluating order dispatch policies on ride-sharing platforms, it is common to randomize the two policies over time. If each hour or half-hour is treated as a single experimental unit -- a standard practice \citep[see e.g.,][Section 5]{shi2023dynamic} -- it yields only a few hundred observations. Second, the signal strength -- defined as the difference in outcomes between the new and standard policies -- is often very small \citep{athey2023semi,sun2024arma}. In practice, improvements from newly-developed order dispatch policies on ride-sharing platforms often range from only 0.5\% to 2\% \citep{tang2019deep}. The third is the carryover effect, which refers to the delayed effect of policies on future outcomes of interest and is ubiquitous in online experiments \citep{xiong2024data}. Consider again the evaluation of order dispatch policies in the motivating ride-sharing application. A dispatch algorithm at a given time not only matches drivers with passengers, directly affecting immediate GMV, but also impacts future GMV by altering the spatial distribution of drivers. Specifically, assigning a driver to an order repositions the driver to the order's destination, changing the locations of drivers across the city, which in turn affects the future GMV (see Figure \ref{fig:carryover} for a visualization).

\subsection{Related works}
There is a growing literature on A/B testing \citep[see][for  recent reviews]{larsen2024statistical,quin2024b}. The core idea behind A/B testing is to leverage causal inference methods to estimate and infer the treatment effect of the new policy. Traditional A/B testing methods are designed for independent and identically distributed (i.i.d.) observations and typically assume the absence of carryover effects. In the absence of confounders affecting both policy assignment and outcomes, testing  the average treatment effect (ATE) can be cast as a two-sample testing problem, where conventional $z$-tests or $t$-tests based on normal or $t$ approximations \citep{student1908probable,berger2001statistical} have proven to be efficient. 

When confounders are present, it suffices to apply classical causal inference methods that operate under the stable unit treatment value assumption \citep[see e.g.,][and the references therein]{imbens2015causal}. These methods can generally be classified into three types: 
\begin{enumerate}
    \item Imputation methods  \citep[see e.g.,][]{rubin1979using,abadie2011bias,ye2023toward}, which impute missing potential outcomes from regression models;
    \item Weighting methods, particularly inverse propensity score weighting \citep[IPW,][]{Paul1983the,zhou2015coarsened,li2018balancing,yang2018asymptotic}, that apply an inverse probability weight for each subject to obtain consistent estimators;
    \item Augmented IPW (AIPW) methods, as well as their variants, including double machine learning \citep[see e.g.,][]{scharfstein1999adjusting,bang2005doubly,zhang2012robust,athey2018approximate,Chernozhukov2018double,Wang2020debiased}, that combine the virtues of imputation methods and weighting methods to achieve consistency under milder conditions.
\end{enumerate}
The aforementioned three types of methods are also referred to as the direct method, importance sampling (IS) method and doubly robust (DR) method in machine learning \citep{dudik2014doubly,uehara2022review}. 

More recent proposals have focused on dynamic settings, where randomization is conducted over time in the presence of carryover effects. Naturally, existing methods from the causal inference literature designed to handle carryover effects are applicable \citep[see, e.g.,][]{robins1986new, sobel2014causal, bojinov2019time}. In parallel, a growing body of works has proposed to adopt a reinforcement learning \citep[RL,][]{sutton2018reinforcement} framework for A/B testing, by leveraging the widely studied Markov decision process \citep[MDP,][]{puterman2014markov} model in RL to explicitly capture the carryover effect \citep[see e.g.,][]{glynn2020adaptive,farias2022markovian,li2023optimal,shi2023dynamic,chen2024experimenting,wenunraveling2025}. Specifically, under the MDP assumption, existing off-policy evaluation (OPE) methods from the RL literature \citep[see,][for a review]{uehara2022review} can be applied to estimate the expected outcome under each policy -- and thus used to estimate their difference, i.e., the treatment effect. These methods include extensions of imputation, weighting and AIPW methods to the MDP setting \citep[see e.g.,][]{bradtke1996linear,precup2000eligibility,zhang2013robust,thomas2015high,jiang2016doubly,le2019batch,luckett2020estimating,kallus2022efficiently,liao2022batch,shi2022statistical}, as well as model-based approaches that estimate the MDP model from the data to derive the ATE estimator \citep[see e.g.,][]{luo2022policy}. We also note that many of these OPE methods yield asymptotically normal estimators, which can be used as test statistics to assess the statistical significance of the improvement under the new policy \citep[Section 5]{shi2025statistical}.

Finally, other related works have explored (i) sequential monitoring, which conducts A/B testing at multiple interim stages to enable early termination without inflating the overall type-I error \citep{johari2015always,waudby2024anytime}; (ii) the careful design of experiments to enhance the efficiency of ATE estimators \citep{bojinov2023design,xiong2024data,li2023optimal,sun2024arma,wenunraveling2025,ni2025enhancing}; (iii) methods for handling interference effects beyond temporal carryover effects, such as spatial, network, or marketplace interference \citep{ugander2013graph,hu2022average,bajari2021multiple,leung2022rate,viviano2023causal}.

\subsection{Contributions}
\begin{figure}[t]
	\centering
	\begin{minipage}{0.45\textwidth}
		\includegraphics[height=5cm,width=1\textwidth]{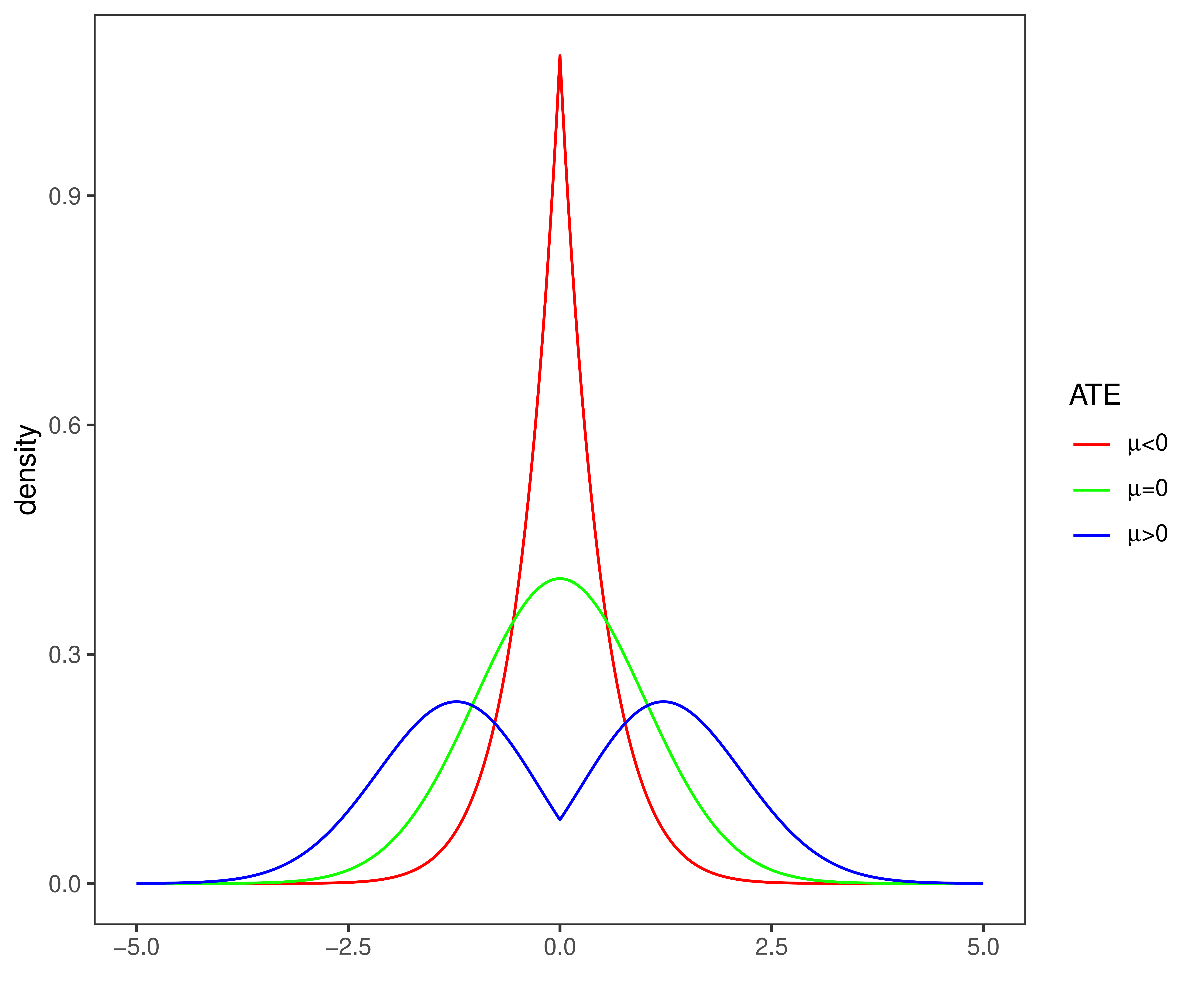}
		\centerline{(a)}
	\end{minipage}	
    \begin{minipage}{0.45\textwidth}
		\includegraphics[height=5cm,width=1\textwidth]{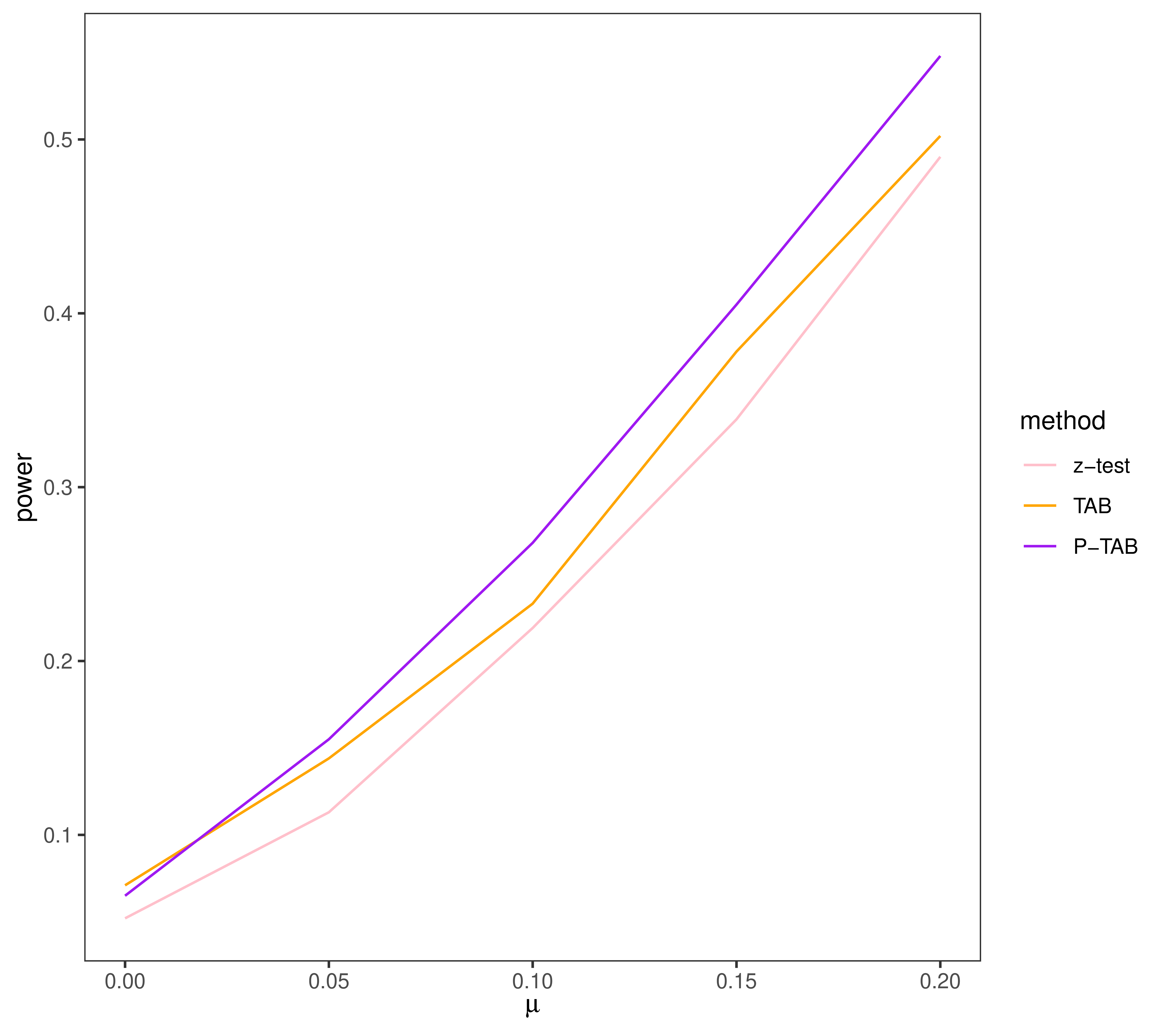}
		\centerline{(b)}
    \end{minipage}
	
	\caption{(a) Probability density functions of bandit distributions under null ($\mu\le 0$) and alternative hypotheses ($\mu>0$). When $\mu=0$, the bandit distribution simplifies to the standard normal distribution. When $\mu<0$, it achieves a more pronounced peak than the standard normal around zero. When $\mu>0$, the distribution becomes bimodal, with less probability concentrated around zero than in the standard normal. (b) Empirical powers of $z$-test, TAB and P-TAB.  TAB achieves higher power than $z$-test by adopting the two-armed bandit framework 
    and P-TAB further improves the power of TAB by employing the permutation procedure.}\label{illustration}
\end{figure}

This paper focuses on A/B testing in both i.i.d. and dynamic settings. 
Our proposal makes useful contributions in the following ways:  
\begin{itemize}[leftmargin=*]
    \item To address the first two challenges, we develop a powerful two-armed bandit (TAB)-based test for inferring the ATE between the new and standard policies by utilizing the recently developed strategic central limit theorem \citep[SCLT,][]{chen2022strategy,Chen2023strategic}. Unlike existing normal-approximation-based test statistics, which primarily differ in their means under the null and alternative hypotheses, the proposed test statistic maintains asymptotically equivalent means while differing in the shape of the distributions under the two hypotheses. This crucial distinction, illustrated in Figure \ref{illustration}(a), results in substantial improvements in statistical power. We further enhance TAB-based test by incorporating a permutation strategy, which reduces sensitivity to sample ordering and significantly boosts the test's power.
    \item To accommodate the last challenge, we extend our proposed test to dynamic settings with carryover effects. 
\end{itemize}

\subsection{Paper organization} 
The rest of the paper is organized as follows. In Section \ref{sec:iid}, we present the proposed test in i.i.d. settings. In Section \ref{sec:dynamic}, we extend the proposed test to dynamic settings. In Section \ref{sec:numericalCase}, we apply the proposed test to five real datasets from a ride-sharing company to evaluate the treatment effects of order dispatch as well as subsidy policies. Finally, Section \ref{sec:discussion} concludes our paper. Technical proofs and additional simulation studies are relegated to the Supplementary Materials.

\section{A/B testing in i.i.d. settings}\label{sec:iid}
This section introduces a two-armed bandit framework for A/B testing in i.i.d. settings. We first introduce our testing hypotheses in Section \ref{subsec:testhyp}. We next present the main idea of our test in Section \ref{subsec:oracle} and detail the implementation in Section \ref{sec:testpermutation}. 
\subsection{Testing hypotheses}\label{subsec:testhyp}
We adopt a potential outcome framework to formulate our testing hypotheses.
Let $X$ be a $p$-dimensional vector capturing a customer's baseline characteristics. Let $A$ represent a binary treatment indicator, where, by convention, 1 stands for the newly-developed policy, and 0 for the standard control. Let $Y$ denote the outcome of interest, where higher values are preferred. Beyond these observed variables, we introduce two potential outcome variables: $Y^{(0)}$ and $Y^{(1)}$, denoting the outcomes the company would achieve if treatment 0 or 1 were employed, respectively.

At the population level, the ATE is defined as the average difference between the two potential outcomes, 
$$\mu \triangleq \textrm{ATE} = \mathbb{E}(Y^{(1)} - Y^{(0)}).$$
Our testing hypotheses are formalized as:
\begin{equation} \label{nullhy}
\mathcal{H}_0: \mu \leq 0  \text{~~v.s.~~} \mathcal{H}_1: \mu >0.
\end{equation}
When the null hypothesis $\mathcal{H}_0$ holds, there is no sufficient evidence to suggest that the new policy is superior to the control. Given the costs associated with implementing a different policy, we recommend to continue with the control.  
Conversely, when the alternative hypothesis $\mathcal{H}_1$ holds, the new policy significantly outperforms the standard control and we recommend to switch to the new policy.

The potential outcomes $Y^{(0)}$ and $Y^{(1)}$ are not identifiable without additional conditions. To consistently infer the ATE, we adopt the following set of conditions, which are frequently imposed in causal inference. 
\begin{assumption}[Consistency]\label{sutva}
The observed outcome is equal to its corresponding potential outcome under the observed treatment, i.e., $Y=Y^{(A)}$, almost surely.
\end{assumption}

\begin{assumption}[Unconfoundedness]\label{ignorability}
Given covariates ${X}$, treatment assignment of $A$ is independent to the potential outcomes, i.e., $A \perp (Y^{(0)}, Y^{(1)})\mid {X}$.
\end{assumption}

\begin{assumption}[Positivity]\label{positivity}
There exists some $\epsilon>0$ such that
$$\mathbb{P}(A=a\mid {X}={x}) > \epsilon,\quad \forall  a \text{~and~} {x}.$$
\end{assumption}

Assumption \ref{sutva}, commonly referred to as the consistency assumption, establishes a crucial connection between observed and potential outcomes. 
Assumption \ref{ignorability}, also known as the ignorability assumption, essentially requires the collection of a sufficient set of baseline covariates to fulfill the conditional independence.
Assumption \ref{positivity}, denoted as positivity, requires the treatment assignment to be non-deterministic for any value of ${X}$. It is also referred to as the overlap condition \citep[see e.g.,][]{kallus2018policy} in machine learning.
The latter two conditions are automatically satisfied in randomized studies. 
It can be shown that under Assumptions \ref{sutva}-\ref{positivity}, the ATE can be re-expressed as follows \citep[see e.g.,][]{rosenbaum1983central,shpitser2010validity},
\begin{equation}\label{eqn:inferATE}
    \mu=\mathbb{E}[\mathbb{E}(Y|A=1,X)-\mathbb{E}(Y|A=0,X)].
\end{equation}
Notice that the right-hand side (RHS) of \eqref{eqn:inferATE} consists solely of observable quantities and does not involve latent potential outcomes. This ensures that the ATE is ``learnable'' from the observed data.



\subsection{Oracle test via two-armed bandit}\label{subsec:oracle}

We begin by introducing the two-armed bandit process, a classical model in the realm of probability theory and decision process \citep[see e.g.,][]{lai1987adaptive}. The two-armed bandit problem can be viewed as perhaps the simplest form of the broader reinforcement learning problem, which has become one of the most popular research topics in machine learning \citep{sutton2018reinforcement}. 
In this problem, an agent faces a binary choice between two policies, referred to as ‘arms’. Each arm yields rewards governed by unknown probability distributions. The agent makes sequential decisions, selecting one arm $\theta_t\in \{0,1\}$ at each time $t$ and subsequently receives a reward $R_t\in \mathbb{R}$. These decisions are guided by the knowledge gained from the observed history, denoted by $H_{t}=\{(\theta_k,R_k):k<t\}$. Formally speaking, at each time $t$, the agent selects the arm according to a mapping $\pi_t$ from $H_{t}$ to a probability mass function on $\{0,1\}$ such that $\mathbb{P}(\theta_t=1|H_t)=\pi_t(H_{t})$. The ultimate goal is to determine the optimal policy $\bar{\pi}_n=\{\pi_t\}_{t=1}^n$, a collection of these mappings, to maximize the expected cumulative reward over time.  



Consider a scenario where $n$ visitors are enrolled in the online experiments, each associated with two potential outcomes, denoted by $Y_i^{(0)}$ and $Y_i^{(1)}$. 
As mentioned in the introduction, in practice, only one of these potential outcomes can be observed per subject. However, to better illustrate the main idea, we assume both outcomes can be observed and develop an ``oracle'' test in this subsection. 
The methodology for addressing missing outcomes will be detailed in the next subsection.

Assume the variance of the difference between the two potential outcomes, denoted by $\sigma^2$, is known.
A commonly used test for assessing \eqref{nullhy} is the $z$-test, whose test statistic can be expressed in one of two forms: 
$$T_n(0) = \sum_{i=1}^n\frac{Y_i^{(1)}-Y_i^{(0)}}{\sqrt{n}\sigma},$$
or
$$T_n(1) = -T_n(0)= \sum_{i=1}^n\frac{Y_i^{(0)}-Y_i^{(1)}}{\sqrt{n}\sigma}.$$
Under the null hypothesis where $\mu \le 0$, 
$T_n(0)$ is asymptotically equivalent or stochastically smaller than 
a standard normal random variable whereas $T_n(1)$ is asymptotically equivalent or stochastically larger than such a variable. Conversely,
under the alternative hypothesis where $\mu>0$, $T_n(0)$ tends toward $+\infty$ and $T_n(1)$ tends toward $-\infty$. Therefore, we reject the null hypothesis when $T_n(0)$ is significantly large or equivalently, when $T_n(1)$ is significantly small. 

To connect the test statistics $T_n(0)$, $T_n(1)$ with the two-armed bandit process, assume the visitors arrive sequentially in order. For each $i$th subject, the agent faces a choice between two arms: selecting the left arm ($\theta_i=0$) results in an immediate reward of $(Y_i^{(1)}-Y_i^{(0)})/\sqrt{n}\sigma$, whereas choosing the right arm ($\theta_i=1$) yields $(Y_i^{(0)}-Y_i^{(1)})/\sqrt{n}\sigma$. Each policy $\bar{\pi}_n$ uniquely determines an action sequence $\{\theta_i\}_i$, leading to a cumulative reward
\begin{equation*}
    T_n(\bar{\pi}_n)=\sum_{i=1}^n \Big[(1-\theta_i)  \frac{Y_i^{(1)}-Y_i^{(0)}}{\sqrt{n}\sigma} +\theta_i \frac{Y_i^{(0)}-Y_i^{(1)}}{\sqrt{n}\sigma}\Big].
\end{equation*}
This cumulative reward serves as our test statistic. Consequently, each policy $\bar{\pi}_n$ effectively determines a test statistic. In the context of hypothesis testing, our objective is not necessarily to identify the optimal policy $\bar{\pi}_n$ that maximizes the expected value of $T_n(\bar{\pi}_n)$, but rather to select a policy that maximizes the power of the resulting test based on $T_n(\bar{\pi}_n)$. 

Under this framework, the two aforementioned $z$-tests operate as follows: the first test consistently chooses the left arm throughout, obtaining a total reward of $T_n(0)$. Conversely, the second test always selects the right arm, resulting in the total reward of $T_n(1)$. These non-dynamic policies are specifically chosen to maximize the power of the tests within their respective classes, as we detail below.

We classify all tests based on their rejection regions. Consider the following two classes of one-tailed tests:
\begin{itemize}
    \item \textbf{Class I} tests $T_n(\bar{\pi}_n)$, where the rejection region is defined as $(C(\bar{\pi}_n), +\infty)$;
    \item \textbf{Class II} tests $T_n(\bar{\pi}_n)$, where the rejection region is defined as $(-\infty, C(\bar{\pi}_n))$.
\end{itemize}
For both classes, $C(\bar{\pi}_n)$ is calculated such that the probability of falling into the resulting rejection region under the null is bounded by a specified significance level $\alpha>0$. In particular, when $\mu=0$, for each $\bar{\pi}_n$, the test statistic $T(\bar{\pi}_n)$ corresponds to the sum of a marginal difference sequence, following a standard normal distribution asymptotically \citep[see e.g.,][]{hall2014martingale}. Consequently, $C(\bar{\pi}_n)$ for the two classes equals the $1-\alpha$th and $\alpha$th quantiles of a standard normal random variable, denoted by $z_{1-\alpha}$ and $z_{\alpha}$, respectively. Note that these critical values are independent of $\bar{\pi}_n$. 

Consequently, it suffices to identify the optimal in-class policies that maximizes $\mathbb{P}(T_n(\bar{\pi}_n)>z_{1-\alpha})$ or $\mathbb{P}(T_n(\bar{\pi}_n)<z_{\alpha})$ under the alternative hypothesis, for their respective policy classes. It becomes evident that the two non-dynamic policies employed by the conventional $z$-tests are strategically chosen to maximize their powers under the alternative hypothesis. This rationale supports the use of $z$-tests under the two-armed bandit framework. 

Next, we introduce a third class of tests:
\begin{itemize}
    \item \textbf{Class III} tests, where the rejection region is defined as $(-\infty, -C(\bar{\pi}_n))\cup (C(\bar{\pi}_n), +\infty)$.
\end{itemize}
Different from Classes I and II, this class constructs \textit{two-tailed} tests for \textit{one-sided} hypothesis in \eqref{nullhy}. Although this may initially seem counterintuitive, the benefit of using these two-tailed tests is that they enable the optimal in-class policy to be dynamic, meaning it 
will not consistently favor either the left or the right arm. 
Similarly, by restricting to the null hypothesis where $\mu=0$, $T(\bar{\pi}_n)$ is asymptotically standard normal, and thus $C(\bar{\pi}_n)=z_{1-\alpha/2}$, being independent of $\bar{\pi}_n$. 

Consider the following dynamic policy $\bar{\pi}_n^*=\{\pi_t^*\}_{t=1}^n$ such that $\pi_1^*$ uniformly randomly selects an action, i.e., $\pi_1^*(H_1)=0.5$, and 
\begin{equation}\label{eqn:strategy}
\pi_t^*(H_t) = \bigg\{ 
\begin{array}{cl}
0, & T_{t-1}(\bar{\pi}_{t-1}^*) >0, \\
1, & T_{t-1}(\bar{\pi}_{t-1}^*)\leq 0.
\end{array} 
\end{equation}
According to SCLT \citep[e.g.,][Theorem 3.3]{chen2022strategy}, $T_n(\bar{\pi}_n^*)$ follows a bandit distribution asymptotically and satisfies
\begin{enumerate}
    \item[(i)] $\lim_n P(|T_n(\bar{\pi}_n^*)|> z_{1-\alpha/2}|\mathcal{H}_0) \leq \alpha$ for any choice of $\mu\le 0$,  ensuring that the test based on $\bar{\pi}_n^*$ controls the type-I error.
    \item[(ii)] $\lim_n P(|T_n(\bar{\pi}_n^*)| > z_{1-\alpha/2}|\mathcal{H}_1) = \lim_n \mathop{\max}\limits_{\bar{\pi}_n} P(|T_n(\bar{\pi}_n)| > z_{1-\alpha/2}|\mathcal{H}_1)$, indicating that $\bar{\pi}_n^*$ is indeed the optimal policy that maximizes the power of class III tests.
\end{enumerate}

This yields the TAB-based test under an oracle condition with observable potential outcomes, where we reject the null hypothesis if $|T_n(\bar{\pi}_n^*)|>z_{1-\alpha/2}$. Its associated $p$-value is given by $2\Phi(-|T_n(\bar{\pi}_n^*)|)$, where $\Phi(\bullet)$ denotes the cumulative distribution function of a standard normal random variable. 
To demonstrate why this test is powerful, Figure \ref{illustration}(a) visualizes the probability density function (pdf) of bandit distribution -- the asymptotic distribution of $T_n(\bar{\pi}_n^*)$, which can be expressed as
\begin{equation}\label{eqn:banditdensity}
    f(y|\kappa_n,\sigma_0)=\frac{1}{\sqrt{2\pi}\sigma_0}\exp\Big(-\frac{(|y|-\sigma_0*\kappa_n)^2}{2\sigma_0^2}\Big)- \frac{\kappa_n}{\sigma_0} \exp\Big(\frac{2\kappa_n|y|}{\sigma_0}\Big) \Phi\Big(-\frac{|y|}{\sigma_0}-\kappa_n\Big),
\end{equation}
with $\kappa_n$ being $\sqrt{n} \mu/\sigma$ and $\sigma_0$ being $\sqrt{1+\mu^2/\sigma^2}$ \citep{Chen2023strategic}. To conclude this section, we discuss two aspects of this distribution: its center and shape. 

\smallskip

\noindent \textbf{Center}. It can be seen from Figure \ref{illustration}(a) that the tests under both null and alternative hypotheses are symmetric around their center, which is zero. This is also evident from the form of pdf in \eqref{eqn:banditdensity}. It implies that the asymptotic mean of the TAB-based test statistic $T_n(\bar{\pi}_n^*)$ remains unchanged when transitioning from the null to the alternative hypothesis. 
The basis for this symmetry lies in the use of the dynamic policy detailed in \eqref{eqn:strategy}. Since the final test statistic's value inversely depends on the initial policy choice — being negative if the initial policy selects the right arm compared to the left — the statistic remains symmetric around zero, due to that both arms are selected with equal probability initially.

\smallskip

\noindent \textbf{Shape}. In contrast to its center, the shape of the bandit distribution varies substantially under the null and alternative hypotheses:
\begin{itemize}
    \item When $\mu=0$, as depicted in green and evident from \eqref{eqn:banditdensity}, the bandit distribution simplifies to the standard normal;
    \item When $\mu<0$, as depicted in red, the bandit distribution achieves a more pronounced peak compared to the standard normal, with a greater concentration of probability near zero. This behavior can be theoretically verified according to \eqref{eqn:banditdensity}. When $y=0$, its derivative with respect to $\kappa_n$ equals $-\Phi(-\kappa_n)$, which is negative. This implies $f(0|\kappa_n,\sigma_0)$ is monotonically decreasing as a function of $\kappa_n$. Consequently, the bandit distribution achieves a higher density at zero when $\kappa_n=\sqrt{n}\mu/\sigma<0$, or equivalently, when $\mu<0$. 
    \item When $\mu>0$, the distribution becomes bimodal, with two peaks distanced from zero and less probability concentrated around the mean than in the standard normal. A closer examination at \eqref{eqn:banditdensity} reveals that the two peaks are centered around  $\pm \kappa_n\sigma_0=\pm \sqrt{n}\mu \sqrt{1+\mu^2/\sigma^2}/\sigma$, respectively. By definition, these peaks diverge to infinity as $n$ increases. 
\end{itemize}
This difference in the shape allows us to distinguish between the null and alternative hypotheses. Specifically, under the null ($\mu\le 0$), the bulk of the test distribution is centered around zero, whereas under the alternative ($\mu>0$), the bulk shifts to the two peaks, away from zero. Consequently, the absolute value of the test statistic is informative in making the decision on whether to reject the null hypothesis, leading to the rejection region: $|T_n(\bar{\pi}_n^*)|>z_{1-\alpha/2}$. As shown in Figure \ref{illustration}(b), the resulting test is more powerful than the conventional $z$-test. 

\subsection{Practical test via permutation and pseudo outcome construction}\label{sec:testpermutation}
The TAB-based test statistic discussed in Section \ref{subsec:oracle} possesses a well-defined limiting distribution and demonstrates favorable power properties. However, it suffers from two limitations: 
\begin{enumerate}
    \item Unlike the $z$- or $t$-test, constructing this test depends on the ordering of the samples and is not ordering insensitive, leading to the ``$p$-value lottery" \citep{meinshausen2009p}.
    \item It requires both potential outcomes to be observable, which is infeasible in practice as only one of them can be observed.
\end{enumerate}
To address the first limitation, we mitigate ordering sensitivity by randomly permuting the samples multiple times, with each permutation leading to a test, and aggregate all these tests to produce a composite statistic. 
To address the second limitation, we employ doubly robust estimation to construct pseudo outcomes that approximate the difference between the two potential outcomes in the randomized control trails, and the method to deal with A/B testing procedure is detailed later.

\smallskip

\noindent \textbf{Permutation}. The optimal policy $\bar{\pi}_n^*$ specified in \eqref{eqn:strategy} is ordering sensitive, which results in the test statistic $T_n(\bar{\pi}_n^*)$ also being sensitive to ordering. In other words, each ordering can yield a potentially different $T_n(\bar{\pi}_n^*)$, although these test statistics are asymptotically equivalent. The test discussed in Section \ref{subsec:oracle} can be viewed as randomly picking one of these $T_n(\bar{\pi}_n^*)$ values. This inherent randomness introduces additional variability into the test, reducing its power in finite samples.

We employ a permutation-based approach to enhance ordering robustness and improve the power. More specifically, we randomly generate $B>1$ many permutations, each being a function $\Pi_b$ that maps a particular subject $i\in \{1, \dots, n\}$ to $\{1, \dots, n\}$ such that $\Pi_b(i_1)\neq \Pi_b(i_2)$ whenever $i_1\neq i_2$. For $b=1,\dots, B$, 
we apply $\Pi_b$ to the $n$ potential outcomes to obtain a permutated sample $\{(Y_{\Pi_b(i)}^{(0)}, Y_{\Pi_b(i)}^{(1)}): 1\le i\le n\}$, apply the optimal policy
$\bar{\pi}_n^*$ to this permutated sample 
to construct the permutated statistic $T_n^{(b)}(\bar{\pi}_n^*)$, calculate its $p$-value 
$p_b = 2 \Phi( - |T_n^{(b)}(\bar{\pi}_n^*)| )$, and employ a $p$-value combination method to aggregate all these $p$-values to produce the final test statistic. 

There exist various 
$p$-value aggregation methods, such as the normal distribution-based method \citep{hartung1999note}, the quantile-based method \citep{meinshausen2009p}, and the Cauchy combination method \citep{Liu2020cauchy}, to mention a few. 
For instance, 
the quantile-based method 
aggregates all $p$-values using their empirical quantile, given by
$$Q(\gamma) = \min \{1, q_{\gamma}(\{p_b/\gamma, b=1,\cdots, B\})\},$$
where $\gamma \in (0,1)$ denotes a pre-specified quantile level, and $q_\gamma(\cdot)$ denotes the empirical $\gamma$th upper quantile. It can be shown that when each $p_b$ is a valid $p$-value, then their empirical quantile $Q(\gamma)$ is also valid.

The Cauchy combination method 
aggregates all these individual $p$-values as follows, 
\begin{equation}\label{eqn:cauchycombine}
    \widetilde{T}_n=\frac{1}{B} \sum_{b=1}^B \tan[(0.5-p_b)\pi],
\end{equation}
where tan denotes the tangent function. To illustrate the rationale behind the Cauchy combination, consider the null hypothesis where $\mu=0$. In this case, all test statistics $T_n^{(b)}(\bar{\pi}_n^*)$s across different permutations are asymptotically normal. Hence, their $p$-values, calculated as $2 \Phi(-|T_n^{(b)}(\bar{\pi}_n^*)| )$, are uniformly distributed between 0 and 1. In the two extreme scenarios where the individual $p$-values are either (i) completely independent or (ii) completely identical, $\widetilde{T}_n$ in \eqref{eqn:cauchycombine} follows a standard Cauchy distribution. In more general scenarios that lie in the middle between these two extremes,  \cite{Liu2020cauchy} showed that the tail of \eqref{eqn:cauchycombine} can still be well approximated by a standard Cauchy distribution. Therefore, for a given $\widetilde{T}_n$, its $p$-value can be calculated as $0.5 - \arctan(\widetilde{T}_n)/\pi$, and we reject the null if the $p$-value is smaller than the significance level $\alpha$. 

\begin{remark}
Compared with classic approaches for combining $p$-values, such as Fisher's method \citep{fisher1928statistical}, the
quantile-based and Cauchy combination methods accommodate a wider range of dependency structures among $p$-values and offer an analytically derived expression for the final $p$-value. They have also been widely employed in practice \citep[see e.g.,][]{mccaw2020operating,shi2022testing,chen2023multi}. 
\end{remark}

Since the aggregated test is constructed using permutations under the two-armed bandit framework, we refer it to as P-TAB. As illustrated in Figure \ref{illustration}(b), P-TAB, when coupled with the Cauchy combination, further enhances the power of the original TAB-based test in finite samples.

\smallskip

\noindent \textbf{Pseudo outcome construction}. In practice, the potential outcomes $Y^{(0)}$ and $Y^{(1)}$ cannot be fully observed. 
In this subsection, we construct pseudo outcomes as surrogates to derive the test.

For a given covariates-policy-outcome triplet $(X,A,Y)$, 
let $m(a,x)$ and $b(a,x)$ denote the outcome regression function and the propensity score function such that $m(a,x)=\mathbb{E} (Y|A=a,X=x)$ and $b(a,x)=\mathbb{P}(A=a|X=x)$. Our approach applies (A)IPW to these observed triplets to approximate the pseudo outcomes. Specifically, according to \eqref{eqn:inferATE}, the ATE can be expressed as the average difference between the two conditional expectations. IPW is motivated by the change of measure theorem, which shows that each averaged conditional expectation can be expressed as follows: 
\begin{equation*}
    \mathbb{E}[\mathbb{E}(Y|A=a,X)]=\mathbb{E} \Big[\frac{\mathbb{I}(A=a)}{b(a,X)}Y\Big].
\end{equation*}
Notice that the RHS essentially corresponds to a weighted average of the observed outcome with weight being the importance sampling (IS) ratio -- 
used to adjust the distributional shift between the target treatment and the treatment assignment mechanism in the observed data. As such, the following pseudo outcome is unbiased to the ATE,
\begin{equation}\label{eqn:some}
    \Big[\frac{\mathbb{I}(A_i=1)}{b(1,X_i)}-\frac{\mathbb{I}(A_i=0)}{b(0,X_i)}\Big]Y_i.
\end{equation}
In addition, AIPW can be further employed to mitigate the variance of \eqref{eqn:some} arisen from the use of the IS ratio. This adjustment yields the following pseudo outcome,
\begin{equation}\label{eqn:someother}
    \widehat{\mu}_i=m(1,X_i)-m(0,X_i)+\frac{\mathbb{I}(A_i=1)}{b(1,X_i)} [Y_i-m(A_i,X_i)]-\frac{\mathbb{I}(A_i=0)}{b(0,X_i)} [Y_i-m(A_i,X_i)].
\end{equation}
To understand the connection between \eqref{eqn:some} and \eqref{eqn:someother}, notice that when $m\equiv 0$, $\widehat{\mu}_i$ is reduced to the IPW-based pseudo outcome in \eqref{eqn:some}. More generally, $\widehat{\mu}_i$ achieves the same expected value as \eqref{eqn:some} provided that the propensity score $b$ is correctly specified, regardless of the correctness of $m$. 

However, the pseudo outcome in \eqref{eqn:someother} offers two advantages over the one in \eqref{eqn:some}:
\begin{enumerate}
    \item \eqref{eqn:someother} generally achieves a smaller variance when compared to \eqref{eqn:some}. Specifically, a well-specified model for $m$ can significantly reduce the variance of $\widehat{\mu}_i$. In fact, the variance of $\widehat{\mu}_i$ is minimized when $m$ is correctly specified \citep{tsiatis2006semiparametric}.
    \item \eqref{eqn:someother} requires a weaker condition than \eqref{eqn:some}. While \eqref{eqn:some} requires the correct specification of $b$ to achieve unbiasedness to $\mu$, $\widehat{\mu}_i$ in \eqref{eqn:someother} is unbiased when either $b$ or $m$ is correctly specified, a characteristic known as the doubly robust property.
\end{enumerate}
It remains to estimate the nuisance functions $m$ and $b$ to construct the pseudo outcomes $\{\widehat{\mu}_i\}_i$. These functions can be estimated using state-of-the-art nonparametric regression or machine learning algorithms. Even if these estimators converge at a rate slower than the root-\( n \) rate, the resulting test remains theoretically sound, as discussed in Section 1.3
of the Supplementary Materials. 

Specifically, when auxiliary datasets are available, they can be utilized to estimate the two nuisance functions, which are then plugged into \(\eqref{eqn:someother}\) to construct the pseudo outcome.
Alternatively, sample-splitting and cross-fitting can be employed \citep{Chernozhukov2018double}. This method divides the data into \( K \) non-overlapping subsets, \(\cup_k \mathcal{D}_k\), each of equal size. For each \( k \), we estimate \( m \) and \( b \) using all data excluding \(\mathcal{D}_k\), and then plug these estimators into \(\eqref{eqn:someother}\) to construct the pseudo outcome \(\widehat{\mu}_i\) for any \( i \) in \(\mathcal{D}_k\). This process is iterated over each \( k \) until the pseudo outcome for each subject is obtained. 

Once we have these \(\widehat{\mu}_i\) values, we use them as surrogates for \( Y_i^{(1)} - Y_i^{(0)} \). The variance term, \( \sigma^2 \), can be estimated using the sampling variance formula, given by $\widehat{\sigma}^2=\sum_i (\widehat{\mu}_i-\overline{\mu})^2/(n-1)$ where $\overline{\mu}=\sum_i \widehat{\mu}_i/n$. This yields the following immediate reward $\pm \widehat{\mu}_i/(\sqrt{n}\widehat{\sigma})$ for each subject $i$. 
\begin{comment}
    
To enhance the finite sample performance, following \citet{Chen2023strategic}, we also include an additional term, $\pm \widehat{\mu}_i/n$ in the calculation of immediate rewards. This adjustment results in the following immediate reward:
\begin{equation*}
    \pm \Big(\frac{\widehat{\mu}_i}{n}+ \frac{\widehat{\mu}_i}{\sqrt{n}\widehat{\sigma}} \Big),
\end{equation*}
for each subject $i$. 
Notice that the additional terms being included $\{\widehat{\mu}_i/n\}_i$, which are of the order $O(n^{-1})$, do not affect the asymptotic performance of the test. However, they were used in the original proposals by \citet{chen2022strategy, Chen2023strategic}, and have been shown to improve the empirical performance of the resulting tests. 
\end{comment}
Finally, we apply P-TAB to these immediate rewards to compute the \( p \)-value. A pseudocode summarizing the proposed method is given in Algorithm \ref{alg}. 
Its theoretical properties are presented below. 

\begin{algorithm}[t]\label{alg:p-tab}
    \caption{P-TAB for ATE testing}
    \label{alg}
        \KwData{$\mathcal{D} = \{(X_i, A_i, Y_i), i=1,\ldots,n\}$}
        \KwResult{$p$-value}

        Divide the data into \( K \) non-overlapping subsets, \(\cup_k \mathcal{D}_k= \mathcal{D}\) , each of equal size.

        \While{$k \leq K$}{
            Estimate nuisance functions $m$ and $b$ using nonparametric regression or machine learning algorithms based on data \(\mathcal{D}\setminus\mathcal{D}_k\) and denote them as $\widehat m^{(k)}$ and $\widehat b^{(k)}$;

            Construct pseudo outcome $\widehat{\mu}_i$ for data in $\mathcal{D}_k$ based on (\ref{eqn:someother}) with $m$ and $b$ replaced by $\widehat m^{(k)}$ and $\widehat b^{(k)}$.
            
        }

        Estimate sample variance for the pseudo outcomes as $\widehat{\sigma}^2=\sum_i (\widehat{\mu}_i-\overline{\mu})^2/(n-1)$ where $\overline{\mu}=\sum_i \widehat{\mu}_i/n$;

        \While{$b \leq B$}{
        Conduct a permutation map $\Pi_p$ to the constructed pseudo outcomes $\{\widehat{\mu}_i, i=1,\ldots, n\}$ to obtain a permutated sample $\{\widehat{\mu}_{\Pi_b(i)}, i=1,\ldots,n\}$;

        Apply the dynamic policy $\bar{\pi}_n^*$ in (\ref{eqn:strategy})  to the permutated sample $\{\widehat{\mu}_{\Pi_b(i)}, i=1,\ldots,n\}$ by defining the rewards as $\frac{\widehat{\mu}_{\Pi_b(i)}}{\sqrt{n}\widehat{\sigma}}$ for the left arm and $-\frac{\widehat{\mu}_{\Pi_b(i)}}{\sqrt{n}\widehat{\sigma}}$ for the right arm to calculate the statistic $T_n^{(b)}(\bar{\pi}^*)$  and its $p$-value $p_b = 2\Phi(-|T_n^{(b)}(\bar{\pi}^*)|)$.
        }

        Aggregate all these $p_b$s using a $p$-value aggregation method (e.g., (\ref{eqn:cauchycombine})) to output the final $p$-value.

\end{algorithm}

\begin{theorem}\label{theory_oracle}
    Under Assumptions \ref{sutva}-\ref{positivity} and Assumptions  \ref{supp-assump:nuisancefunction} in Section \ref{supp-subsec:theoremIID} 
    of the Supplementary Materials, the $p$-value of the proposed permutation- and pseudo-outcome-based two-armed bandit test, denoted by $\widehat{p}$, attains the following properties: 
    \begin{enumerate}
    \item[(i)] \textbf{Type-I error control}: Under the null hypothesis, 
    $$
    \begin{array}{cl}
       \lim_n P(\widehat{p}< \alpha) 
         \leq \alpha.
    \end{array}  
    $$
    \item[(ii)] \textbf{Consistency against fixed alternatives}:  
    For a given fixed $\mu>0$, 
    $$
    \begin{array}{cl}
       \lim_n P(\widehat{p} <\alpha)  
       =1.
    \end{array}
    $$

\end{enumerate}
\end{theorem}
Theorem \ref{theory_oracle} shows that 
pseudo-outcome-based two-armed-bandit test controls the type-I error and remains consistent against alternative hypotheses. This formally establishes its validity and effectiveness.

\section{A/B testing in dynamic settings}\label{sec:dynamic}
This section extends the proposed test to dynamic settings. 
Suppose a technology company is conducting an online experiment 
to assess the efficacy of a newly-developed policy in comparison to a baseline policy. Assume the experiment lasts for $n$ days, and each day is partitioned into $T$ non-overlapping time intervals. Within each day, the data collected from the experiment can be summarized into a trajectory $\{(X_t,A_t,Y_t):1\leq t\leq T\}$. Here, $X_t$ denotes certain market features observed at the beginning of the $t$th time interval, such as the number of call orders and drivers' online time in a ride-sharing platform. $A_t\in \{0,1\}$ denotes the policy the company implemented during the $t$th time interval. $Y_t$ represents the immediate outcome observed (e.g., the total revenue) at the end of the $t$th interval. 

Denote by $\{(X_{i,t}, A_{i,t}, Y_{i,t}):1\leq t\leq T\}$ the trajectory collected at the $i$th day. We assume these trajectories are i.i.d. realizations of $\{(X_t,A_t,Y_t):1\leq t\leq T\}$. We make two remarks. First, the i.i.d. assumption applies across days (trajectories), but not temporally within a single trajectory. This allows for time-dependent observations and carryover effects within each daily trajectory, which are commonly observed in dynamic settings. Second, such an i.i.d. trajectories assumption is mild and likely to hold in various applications such as ride-sharing \citep{li2023optimal,luo2022policy,wenunraveling2025,zhubalancing2025} and marketing auctions \citep{basse2016randomization,liu2020trustworthy}. Take ride-sharing as an example. As shown in Figure~\ref{fig:drivertotalincome}, the number of call orders is very small between 1 a.m. and 5 a.m., which effectively resets the marketplace each day and supports the plausibility of the independence assumption across days. Moreover, the observed variables exhibit consistent patterns across different days, typically peaking during rush hours, which makes the identical distribution assumption reasonable.

Based on the data trajectories collected from the online experiment, our goal is to infer the ATE, defined as
$$\mu=\mathbb{E}^1\Big(\frac{1}{T}\sum_{t=1}^TY_{t}\Big)-\mathbb{E}^{0} \Big(\frac{1}{T}\sum_{t=1}^TY_{t}\Big),$$
where $\E^1$ and $\E^0$ denote the expectations where the new policy (represented by 1) and the baseline policy (represented by 0) are applied across all time intervals, respectively. Similar to \eqref{nullhy}, we wish to test
\begin{eqnarray}\label{eqn:test0}
\mathcal{H}_0: \mu \leq 0  \text{~~v.s.~~} \mathcal{H}_1: \mu >0.  
\end{eqnarray}
Toward that end, we adopt the RL framework that models the trajectory data using an MDP. Specifically, we impose the following Markov assumption. 
\begin{assumption}[Markov assumption]\label{ass:Markov}
The market features and the expected outcomes are assumed to satisfy the Markov property. Specifically, in the case where the market features are discrete,
\begin{eqnarray}\label{eqn:markovstate}
    \mathbb{P}\Big(X_{t+1}=x^{\prime} \mid A_t=a, X_t=x,\{X_j, A_j\}_{j<t}\Big)=\mathbb{P}_t(X_{t+1}=x^{\prime} \mid A_t=a, X_t=x),
\end{eqnarray}
for any $x, a, x^{\prime}$ and $t$. As for the outcomes, we have 
\begin{eqnarray}\label{eqn:markovreward}
    \mathbb{E}\left(Y_t \mid A_t=a, X_t=x,\left\{X_j, A_j\right\}_{j<t}\right)=r_t(a, x),
\end{eqnarray}
for some reward function $r_t$. 
\end{assumption}
Assumptions \eqref{eqn:markovstate} and \eqref{eqn:markovreward}  essentially require that the future market features and expected outcomes are conditionally independent of past market features and policies, given the current market feature and policy. 
We remark that conditions similar to Assumption \ref{ass:Markov} are frequently imposed in the RL literature \citep{sutton2018reinforcement,shi2022statistical,ramprasad2023online}\footnote{Note that Assumptions~\eqref{eqn:markovstate} and \eqref{eqn:markovreward} are strictly weaker than the Markov and conditional mean independence assumptions in \citet{shi2022statistical}. Specifically, unlike the assumptions in \citet{shi2022statistical}, the conditioning sets in \eqref{eqn:markovstate} and \eqref{eqn:markovreward} do not include past outcomes. This difference arises because, in our setting, the number of days $n$ is assumed to grow to infinity, whereas in \citet{shi2022statistical}, $n$ can be finite. In the latter case, consistency requires to incorporate past outcomes into the conditioning sets along with certain mixing conditions.}. 

To apply the proposed TAB-based procedure for testing \eqref{eqn:test0}, we first construct pseudo-outcomes for estimating the ATE. Under Assumption~\ref{ass:Markov}, we adopt the double reinforcement learning estimator \citep[DRL,][]{kallus2020double, liao2022batch} for this purpose. See also Section 4 of \citet{li2023optimal} and Section~4.1 of \citet{wenunraveling2025} for the specific form of the estimator in the context of A/B testing. Specifically, we define 
\begin{equation}\label{eqn:hat_mu_t}
\begin{aligned}
    &\widehat{\mu}_i 
    = \frac{1}{T}\Big[\widehat{V}_1^1(X_{i,1}) - \widehat{V}_1^0(X_{i,1})\Big]\\
    + & \sum_{k=1}^T \sum_{a=0}^1 \frac{(-1)^{a+1}}{T} \, \widehat{\omega}_{k}^{a}(X_{i,k},A_{i,k}) 
    \big[ Y_{i,k} + \widehat{V}_{k+1}^{a}(X_{i,k+1}) - \widehat{V}_k^{a}(X_{i,k}) \big],
\end{aligned}
\end{equation}
as the pseudo outcome for the ATE constructed using the $i$th day's data trajectory. Here, $\widehat{V}_k^a$ and $\widehat{w}_k^a$ denote the estimators for the value function $V_k^a$ and marginalized IS (MIS) ratio $w_k^a$ at time $k$, defined as
\begin{eqnarray*}
    V^a_t(x) = \mathbb{E}^a\Big(\sum_{k=t}^T Y_{k}|X_t=x\Big)\,\,\hbox{and}\,\,\omega_k^a(x,a') = \frac{p_k^a(x,a')}{p_k^b(x,a')},
\end{eqnarray*}
where $p_k^a$ and $p_k^b$ denote the probability mass functions of $X_k$ and $A_k$ under the target policy -- which deterministically assigns $A_t = a$ at each time $t$ -- and the behavior policy used to assign treatments during the online experiment, respectively. When $X_k$s are continuous, their probability density functions can be used to define $p_k^a$ and $p_k^b$. 

We again, make a few remarks regarding the pseudo outcome in \eqref{eqn:hat_mu_t}. First, \eqref{eqn:hat_mu_t} can be viewed as an extension of \eqref{eqn:someother} to the dynamic setting. Similar to \eqref{eqn:someother}, \eqref{eqn:hat_mu_t} is doubly robust in that $\mathbb{E} (\widehat{\mu}_i)=\mu$ whenever $\{\widehat{V}_k^a\}_{k,a}=\{V_k^a\}_{k,a}$ or $\{\widehat{w}_k^a\}_{k,a}=\{w_k^a\}_{k,a}$. Second, the MIS ratio in Equation~\eqref{eqn:hat_mu_t} may be replaced with the per-decision IS (PDIS) ratio \citep{precup2000eligibility, zhang2013robust, thomas2015high}\footnote{Such a PDIS ratio is also referred to as the sequential IS ratio \citep[see e.g.,][]{zhou2025IS}, borrowing terminology from sequential Monte Carlo methods.}, which is computed as a product of IS ratios over time steps -- unlike the MIS ratio, which involves only the IS ratio at time $t$.  However, the resulting pseudo outcome is known to suffer from the curse of horizon \citep{liu2018breaking}. Its variance will grow exponentially fast with respect to the horizon $T$. Finally, directly averaging the pseudo outcomes across days yields an asymptotically normal estimator \citep{kallus2020double} that can be used to test \eqref{eqn:test0}. Below, we employ the TAB procedure for more powerful testing. 

\begin{comment}
To clarify the novel statistic for dynamic settings, we initially assume that $V^1(X_1) - V^0(X_1)$ is known for each day, and provide procedures to estimate these values later. Following the same identification strategy as in Section \ref{sec:iid} to identify the advantage of new policy over baseline policy, we first establish two test statistics as
$$TS_n(0) = \sum_{i=1}^n\frac{V^1(X_{i,1})-V^0(X_{i,1})}{T\sqrt{n}\bar\sigma}, \text{and~~} TS_n(1) = \sum_{i=1}^n\frac{V^0(X_{i,1})-V^1(X_{i,1})}{T\sqrt{n}\bar\sigma},$$
where $\bar\sigma^2$ is the variance of $V^1(X_{1})-V^0(X_{1})$.
Under the null hypothesis $\mathcal{H}_0: \bar\mu \leq 0$, $TS_n(0)$ converges in distribution to a random variable stochastically dominated by a standard normal, while $TS_n(1)$ converges to one that stochastically dominates a standard normal. Under the alternative $\mathcal{H}_1: \bar\mu > 0$, $TS_n(0) \xrightarrow{p} +\infty$ and $TS_n(1) \xrightarrow{p} -\infty$. 
\end{comment}

Specifically, following the methodology in Section \ref{sec:iid}, we compute the following test statistics,
\begin{equation}
	\begin{aligned}
		\text{TS}_{n}(\bar{\pi}_n^*)&=\sum_{i=1}^n
 \frac{(1-\theta_i)\widehat{\mu}_i}{ \sqrt{n}\widehat{\sigma}} - \sum_{i=1}^n  \frac{\theta_i \widehat{\mu}_i}{\sqrt{n}\widehat{\sigma}},
	\end{aligned}
\end{equation}
where $\widehat{\sigma}^2$ denotes the sampling variance of $\{\widehat{\mu}_i\}_i$ and $\theta_i$s satisfy $\Pr(\theta_i=1|H_i)=\pi_i(H_i)$ where $\pi_1^*(H_1)=0.5$, 
\begin{equation}\label{eqn:strategy}
	\pi_i^*(H_i)=\begin{cases}
	0, & 	\text{TS}_{i-1}(\bar{\pi}_{i-1}^*) >0 ,\\
	1, & \text{TS}_{i-1}(\bar{\pi}_{i-1}^*) \leq 0.
	\end{cases}
\end{equation}
This yields the $p$-value $2\Phi(-|\text{TS}_{n}(\bar{\pi}_n^*)|)$. 
Finally, we employ the permutation-based approach to generate multiple $p$-values and combine them to derive the final $p$-value $\widehat{p}_{drl}$. A pseudocode summary of the resulting test is presented in Algorithm~\ref{alg:p-tab-DR}. Its type-I error and power properties are studied in Theorem \ref{theory_drl} below.
\begin{theorem}\label{theory_drl}
     Suppose Assumption \ref{ass:Markov}, and Assumptions
     \ref{assump:bounded}-\ref{assump:basis} of Section \ref{subsec:assumption}
     in the Supplementary Materials hold. Then we have: 
    \begin{enumerate}
    \item[(i)] \textbf{Type-I error control}: Under the null hypothesis, 
    $$
    \begin{array}{cl}
       \lim_n P(\widehat{p}_{drl}< \alpha)
         \leq \alpha.
    \end{array}  
    $$
   
    \item[(ii)] \textbf{Consistency against fixed alternatives}:  
    For a given fixed $\mu>0$, 
    $$
    \begin{array}{cl}
       \lim_n P(\widehat{p}_{drl} <\alpha) 
       =1.
    \end{array}
    $$   
\end{enumerate}
\end{theorem}

\begin{comment}
As the Two-armed Bandit Framework in Section \ref{sec:prob_TAB}, the arm $\theta_i \in \{0,1\}$, for each decision stage  $R_i \in \{ \mu_i^{+}/\sqrt{n}\sigma, \mu_i^{-}/ \sqrt{n}\sigma  \}$, and history $H_i=\{(\theta_k, R_k): k<i\}$, $\Pr(\theta_i=1|H_i)=\pi_i(H_i)$. The ultimate goal is
to determine the optimal policy $\bar{\pi}_n=\{\pi_i\}_{i=1}^n$, a collection of these mappings, to maximize the
expected cumulative reward over time. The optimal policy can be formulated as $\bar{\pi}_n^*=\{ \pi_i^* \}_{i=1}^n$, and $\pi_1^*(H_1)=0.5$, 
\begin{equation}\label{eqn:strategy}
	\pi_i^*(H_i)=\begin{cases}
	0, & 	\text{TS}_{i-1}(\bar{\pi}_{i-1}) >0 ,\\
	1, & \text{TS}_{i-1}(\bar{\pi}_{i-1}) \leq 0 .
	\end{cases}
\end{equation}
    
\end{comment}

\begin{algorithm}[t]
	\caption{P-TAB for ATE testing in order dispatch situations}
\label{alg:p-tab-DR}
	\KwData{$\mathcal{D} = \{(X_{i,t}, A_{i,t}, Y_{i,t}), i=1,\ldots,n; t=1, \cdots, T\}$}
	\KwResult{$p$-value}
	
	Divide the data into \( K \) non-overlapping subsets about $i \in [n]$, \(\cup_k \mathcal{D}_k= \mathcal{D}\) , each of equal size.
	
	\While{$k \leq K$}{
		Estimate nuisance functions $\{V_t^a\}_{t,a}$ and $\{\omega_t^a\}_{t,a}$ using nonparametric regression or machine learning algorithms based on data \(\mathcal{D}\setminus\mathcal{D}_k\) and denote them as $\{\widehat V_{t}^{a,(k)}\}_{t,a}$ and $\{\widehat \omega_t^{a,(k)}\}_{t,a}$;
		
		Construct pseudo outcome $\widehat{\mu}_i$ for data in $\mathcal{D}_k$ based on (\ref{eqn:hat_mu_t}) with $\{\widehat {V}_t^{a,(k)}\}_{t,a}$ and $\{\widehat \omega_t^{a,(k)}\}_{t,a}$.
		
	}
	
	Estimate sample variance for the pseudo outcomes as $\widehat{\sigma}^2=\sum_i(\widehat{\mu}_i-\overline{\mu})^2/(n-1)$ where $\overline{\mu}=\sum_i \widehat{\mu}_i/n$;
	
	\While{$b \leq B$}{
		Conduct a permutation map $\Pi_p$ to the constructed pseudo outcomes $\{\widehat{\mu}_i, i=1,\ldots, n\}$ to obtain a permutated sample $\{\widehat{\mu}_{\Pi_b(i)}, i=1,\ldots,n\}$;
		
		Apply the dynamic policy $\bar{\pi}_n^*$ in (\ref{eqn:strategy})  to the permutated sample $\{\widehat{\mu}_{\Pi_b(i)}, i=1,\ldots,n\}$ by defining the rewards as $\frac{\widehat{\mu}_{\Pi_b(i)}}{\sqrt{n}\widehat{\sigma}}$ for the left arm and $-\frac{\widehat{\mu}_{\Pi_b(i)}}{\sqrt{n}\widehat{\sigma}} $ for the right arm to calculate the statistic $\widehat{	\text{TS}}_{n}^{(b)}(\bar{\pi}^*)$  and its $p$-value $p_b = 2\Phi(-|\widehat{	\text{TS}}_{n}^{(b)}(\bar{\pi}^*)|)$.
	}
	
	Aggregate all these $p_b$s using a $p$-value aggregation method (e.g.,  Cauchy combination method  ) to output the final $p$-value.
\end{algorithm}
\section{Numerical experiments}\label{sec:numericalCase}
In this section, we conduct extensive numerical experiments to evaluate the finite sample performance of the proposed A/B test, using five real datasets from a world-leading ride-sharing company. We evaluate both order dispatch and subsidy policies. Additional simulation studies are conducted in Section \ref{sec:numerical} of the Supplementary Materials. 

\subsection{Application to the evaluation of subsidy policies}
We apply the proposed test to three datasets from the ride-sharing company to demonstrate its usefulness in evaluating passenger-side subsidy policies. 
The first dataset comes from an A/A experiment where all passengers being involved were exposed to the same subsidy policy. This dataset is used to assess the type-I error control of a test, as the null hypothesis should not be rejected given that both groups received the same policy. 
The last two datasets come from two A/B experiments where the company randomly divided passengers into two groups, each exposed to a particular subsidy policy. After the experiment, we compare the GMVs across the two groups of users to evaluate the effectiveness of the two subsidy policies. The new policies in these experiments are expected to yield larger GMVs compared to the existing ones. For all datasets, we use each passenger's pre-experiment GMV as the covariate. The first dataset includes 20,000 passengers. The second and third datasets consist of 22,336 and 20,000 passengers, respectively, each evenly split between control and treatment groups.

In these experiments, randomization is conducted at the passenger level. Accordingly, we treat each passenger's data as i.i.d. and apply the test procedure described in Section~\ref{sec:iid} for policy evaluation. We report the $p$-values of the proposed P-TAB, its variant TAB without permutation as well as the double machine learning-based $z$-test \citep[denoted by DML]{Chernozhukov2018double} in Table \ref{real-world}. As shown, when applied to the first dataset where the null hypothesis holds, all three tests fail to reject the null hypothesis, confirming their validity. For the second and third datasets, both P-TAB and DML reject the null hypothesis at the 5\% significance level. However, TAB fails to reject the null when applied to the second dataset.  Moreover, the proposed P-TAB test consistently produces smaller $p$-values than the other two methods, suggesting that it offers improved power for detecting the alternative hypothesis.

\begin{table}[t]
  \centering
  \caption{$P$-values from the DML, TAB, and P-TAB tests applied to the real-world datasets. }
  \resizebox{0.6\textwidth}{!}{
  {\tiny
  \begin{tabular}{c|ccc}
    \hline
    \tiny{\diagbox{Dataset}{Statistic}} & P-TAB & TAB & DML \\
    \hline
    \textrm{I} & 0.482 & 0.571 & 0.574 \\
    \textrm{II} & 0.044 & 0.086 & 0.055 \\
    \textrm{III} & 0.023 & 0.041 & 0.027 \\
    \hline
    \end{tabular}
    }
    }
    \label{real-world}
\end{table}

\subsection{Application to the evaluation of order dispatch policies}\label{subsec: real_data_based_exps}
Next, we use two additional datasets to investigate the performance of the proposed test in evaluating order dispatch policies. Both datasets span 40 days and are derived from A/A experiments, in which a single order dispatch policy was consistently applied throughout the experiment. These datasets cannot be directly used to assess the power properties of the tests. Following the bootstrap-based simulation procedure of \citet{li2024combining} and \citet{wenunraveling2025}, we use the wild bootstrap method \citep{wu1986jackknife} to construct two simulation environments. A detailed summary of the procedure is provided in Algorithm~\ref{algo:res_bootstrap} in Section~\ref{algrithombootstrap} of the Supplementary Materials.

Specifically, for the evaluation of different order dispatch policies, randomization is conducted over time. In the first dataset, we set the time unit to 30 minutes, resulting in $T=48$ time intervals per day. In the second dataset, we use one hour as the time unit, yielding $T=24$. We adopt a switchback design in which the assigned treatment alternates at each time step, i.e., $A_{i,t}=1-A_{i,t-1}$ for all $t>1$ and $A_{i,1}=1-A_{i-1,T}$ for all $i>1$, with the initial action $A_{1,1}$ being generated uniformly at random. At each time $t$, $X_t$ consists of the number of call orders and the driver's total online time within the last 30-minute or one-hour time interval. $Y_t$ corresponds to the GMV collected from the $t$th time interval. Both variables are simulated using the wild bootstrap algorithm. We also introduce a parameter $\lambda$, which quantifies the percentage improvement of the new order dispatch policy over the existing one. We consider six values of $\lambda$: $0$, $0.2\%$, $0.4\%$, $1\%$, $2\%$, and $5\%$. When $\lambda=0$, the null hypothesis holds; otherwise, the alternative hypothesis holds. See Section~\ref{algrithombootstrap} of the Supplementary Materials for additional details.

\begin{figure}[t]
	\centering
	\begin{minipage}{0.8\linewidth}
		\centering
		\includegraphics[width=\linewidth]{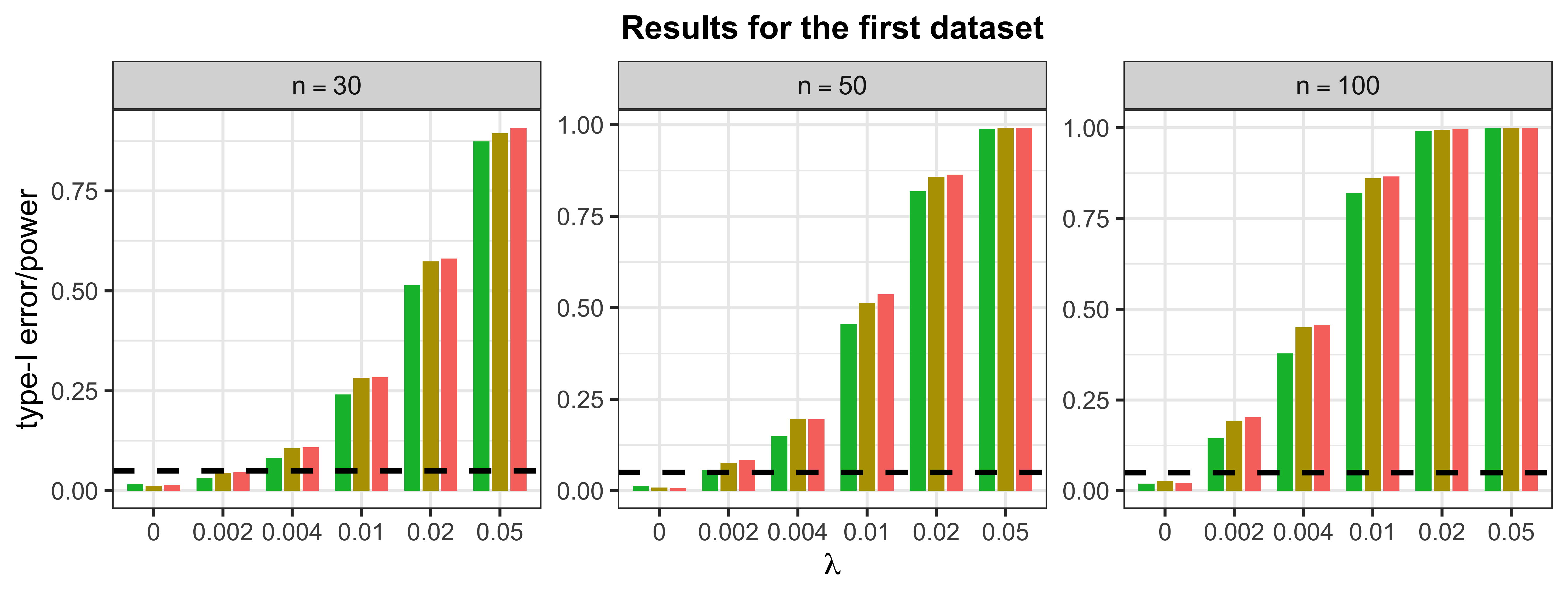}
	\end{minipage}
	\hspace{0.1cm} 
	\begin{minipage}{0.8\linewidth}
		\centering
		\includegraphics[width=\linewidth]{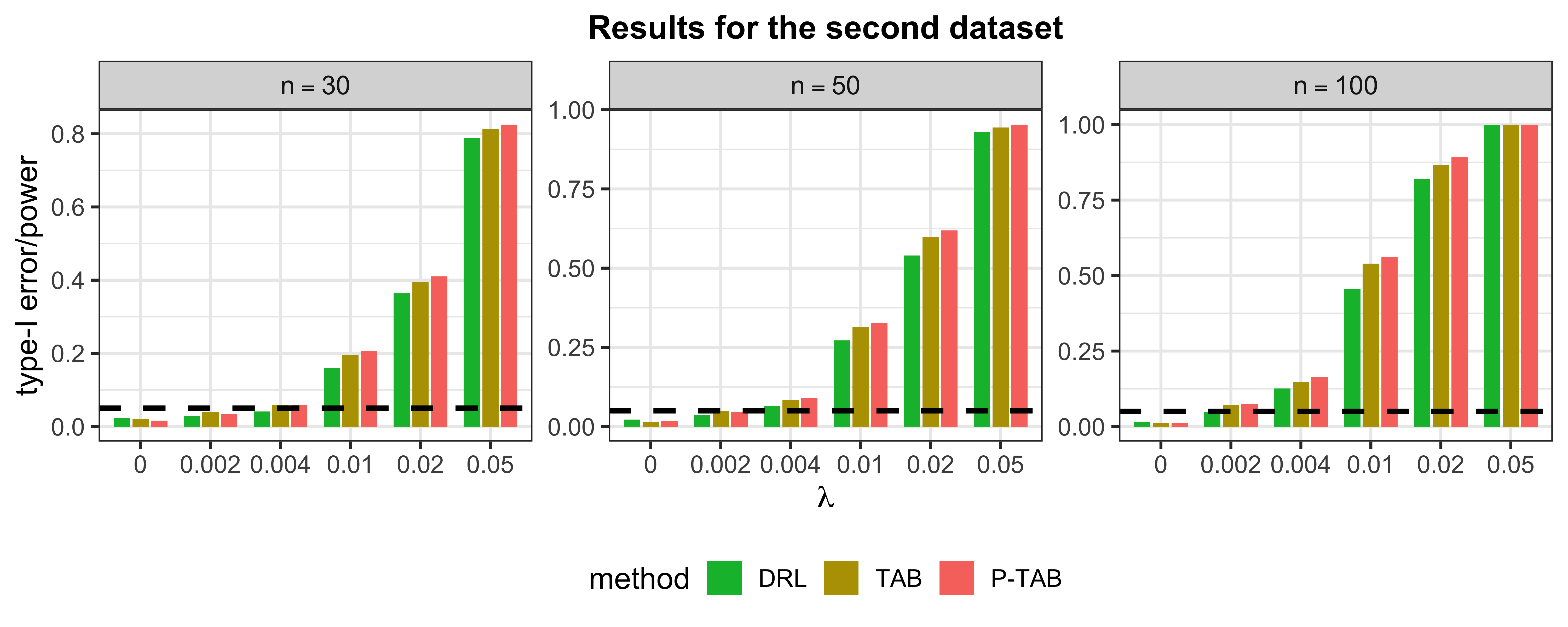}
	\end{minipage}
	
	\caption{\small Type-I errors and powers of different methods under the real data-based environments. Results for $\lambda=0$ indicate type-I errors, and those for $\lambda>0$ indicate powers. }
	\label{fig:real_data_all_probs}
\end{figure}

We apply the proposed P-TAB detailed in Section \ref{sec:dynamic}, its variant TAB and the DRL-based $z$-test -- which calculate the test statistic by taking a simple average over $\widehat{\mu}_i$ (see (\ref{eqn:hat_mu_t})) -- to the simulated data. For each test, we report the proportion of times the null hypothesis is rejected across 1,000 simulation replications in Figure~\ref{fig:real_data_all_probs} and Table~\ref{tab:Real_data} in the Supplementary Materials. These rejection rates correspond to type-I errors in settings where $\lambda=0$, and to powers where $\lambda>0$. It can be seen from Figure~\ref{fig:real_data_all_probs} that 
under the null hypothesis, all three tests control type-I errors below the nominal 0.05 level. Under the alternative hypothesis, P-TAB and TAB achieve higher powers than DRL, with P-TAB outperforming TAB in most scenarios. 


\section{Conclusion}\label{sec:discussion}

In this paper, we present a novel procedure for A/B testing. 
The proposed test contains three key ingredients, including (i) doubly robust  pseudo-outcome estimation; (ii) construction of test statistics within a two-armed bandit framework; and (iii) aggregation of individual $p$-values obtained across multiple permutations. We theoretically establish the validity of the proposed test in terms of type I error control and statistical power. Empirically, we demonstrate its superior power performance over existing methods using five real-world datasets and evaluations of two different types of policies.

\bibliographystyle{imsart-nameyear} 
\bibliography{paperRef}

\begin{center}
{\Large Supplementary Materials}
\end{center}

\setcounter{section}{0} 
\renewcommand{\thesection}{A\arabic{section}}

\setcounter{assumption}{0} 
\renewcommand{\theassumption}{A\arabic{assumption}}

\setcounter{figure}{0} 
\renewcommand{\thefigure}{A\arabic{figure}}

\setcounter{table}{0} 
\renewcommand{\thetable}{A\arabic{table}}

\setcounter{remark}{0} 
\renewcommand{\theremark}{A\arabic{remark}}

 \setcounter{equation}{0}

\section{Theoretical proof}\label{sec:theory}

\subsection{Preliminary}

First we introduce a lemma, which is presented in \cite{Chernozhukov2017double} and \cite{Chernozhukov2018double}.
\begin{lemma}\label{uncondition}
Let $\left\{X_m\right\}$ and $\left\{Y_m\right\}$ be sequences of random vectors. (a) If, for $\epsilon_m \rightarrow 0, \mathbb{P}\left(\left\|X_m\right\|>\epsilon_m \mid Y_m\right) \rightarrow_{P r} 0$, then $\mathbb{P}\left(\left\|X_m\right\|>\epsilon_m\right) \rightarrow 0$. In particular, this occurs if $\mathbb{E}\left[\left\|X_m\right\|^q / \epsilon_m^q \mid Y_m\right] \rightarrow_{P r} 0$ for some $q \geq 1$, by Markov's inequality. (b) Let $\left\{A_m\right\}$ be a sequence of positive constants. If $\left\|X_m\right\|=O_P\left(A_m\right)$ conditional on $Y_m$, namely, that for any $\ell_m \rightarrow \infty$, $\mathbb{P}\left(\left\|X_m\right\|>\ell_m A_m \mid Y_m\right) \rightarrow_{P r} 0$, then $\left\|X_m\right\|=O_P\left(A_m\right)$ unconditionally, namely, that for any $\ell_m \rightarrow \infty$, $\mathbb{P}\left(\left\|X_m\right\|>\ell_m A_m\right) \rightarrow 0$.
\end{lemma}

\subsection{Assumptions} \label{subsec:assumption}

\begin{assumption}[External dataset]\label{supp-assump:externaldata}
    There exists an external dataset with size proportional to $n$ that can be employed to compute the estimated propensity score function $\widehat{b}$ and the outcome regression function $\widehat{m}$.
\end{assumption}
Assumption \ref{supp-assump:externaldata} is primarily imposed to simplify our theoretical analysis. As mentioned in the permutation part, even in the absence of the external dataset, sample-splitting can be employed to eliminate this requirement.
\begin{assumption}[Nuisance functions]\label{supp-assump:nuisancefunction}
(i) $\widehat{b}$ is uniformly bounded away from $0$ and $1$, almost surely; (ii) both $\widehat{m}$ and $m$ are uniformly bounded; (iii) $\sqrt{\mathbb{E} |\widehat{b}(A,X)-b(A,X)|^2}=O(n^{-\ell_1})$ and $\sqrt{\mathbb{E} |\widehat{m}(A,X)-m(A,X)|^2}=O(n^{-\ell_2})$ for some $\ell_1,\ell_2>0$ such that $\ell_1+\ell_2>1/2$. 
\end{assumption}
Assumption \ref{supp-assump:nuisancefunction} is frequently imposed in the literature to establish the asymptotic normality of double machine learning-type estimators \citep[see e.g.,][]{Chernozhukov2018double,diaz2020machine,liang2022estimation}.

\begin{assumption}[Bounded rewards]\label{assump:bounded}
    Rewards $\{Y_t\}$ are uniformly bounded almost surely.
\end{assumption}

\begin{assumption}[External dataset for DRL]\label{supp-assump:ExternalDataII}
There exists an external dataset with size proportional to $n$ that can be employed to compute the estimated marginal density ratios ${\widehat{\omega}}_t^a(X_t,A_t)$s and value functions ${\widehat{V}}_t^a(X_t)$s. 
\end{assumption}

\begin{assumption}[Nuisance functions for DRL]\label{supp-assump:NuisanceII}

The estimated marginal density ratios and value functions satisfy $\mathbb{E}||{\widehat{\omega}}_t^a(X_t,A_t) - \omega_t^a(X_t,A_t)||^2 = o_p(n^{-1/4})$ and $\mathbb{E}||{\widehat{V}}_t^a(X_t) - V_t^a(X_t)||^2 = o_p(n^{-1/4})$ for $1\leq t \leq T$ and $a=0,1$. 
    
\end{assumption}

\begin{assumption}[Basis functions]\label{assump:basis}
Denote the value space for $X_t$s as $\mathcal{X}$.
For each $a \in\{0,1\}, t \in\{1, \ldots, T\}$, there exist $\left\{\theta_{t, a}^*\right\}_{t, a}$ and $\left\{\alpha_{t, a}^*\right\}_{t, a}$ satisfying:
$$
\sup _{\substack{a \in\{0,1\}, x \in \mathcal{X} \\ 1 \leq t \leq T}}\left|V_t^a(x)-\varphi_t^{\top}(x) \theta_{t, a}^*\right|=o\left(n^{-1 / 4}\right) \text { and } \sup _{\substack{a \in\{0,1\}, x \in \mathcal{X} \\ 1 \leq t \leq T}}\left|\omega_t^{a}(x,a)-\varphi_t^{\top}(x) \alpha_{t, a}^*\right|=o\left(n^{-1 / 4}\right) .
$$
    
\end{assumption}

\subsection{Proof of Theorem 1} \label{supp-subsec:theoremIID}

The proofs for Theorem 1 can be structured into three steps.

\noindent{\it Step I}. We first prove that estimated pseudo outcome asymptotically follows the unknown true distribution. Specifically, denote the mean and variance of $\widehat\mu_i$ as $\tilde \mu$ and $\tilde\sigma^2$. We only need to prove
$$\tilde\mu = \mu + o_p(n^{-1/2}), ~~ \tilde\sigma^2 = \sigma^2 + o_p(n^{-1/2}).$$


For notational simplicity, denote $W = (X, A, Y)$ and $\omega := \omega(A,X) = c(m(A, X), b(A, X))$.
First we give some notations. 
Define function
\begin{equation}\label{psi1}
\psi(W; \omega) = m(1, {X}) - m(0,X) + \frac {\mathbb{I}(A=1)}{b({1,X})} [Y - m(A,{X})] - \frac {\mathbb{I}(A=0)}{b({0,X})} [Y - m(A,{X})].
\end{equation}
And denote the score function
$$\Psi(W; \omega) = \psi(W; \omega) - \mu. $$
Then we have 
$$\mathbb{E}(\Psi(W; \omega_0)) = 0,$$
with $\omega_0$ being the true nuisance function.
And the orthogonality condition defined in \cite{Chernozhukov2018double} holds for $\Psi(W; \omega)$.
Note that
$$\tilde\mu - \mu = \mathbb{E}[\psi(W; \widehat{\omega})] - \mathbb{E}[\psi(W; \omega_0)],$$
where $\widehat{\omega}$ indicates that the function is estimated based on the external dataset $\mathcal{D}$.
Given external data, we investigate $\mathbb{E}[\psi(W; \widehat{\omega})| \mathcal{D}] - \mathbb{E}[\psi(W; \omega_0)]$.
Denote $f(t):=\mathbb{E}\left[\psi\left(W ; \omega_0+t\left(\omega-\omega_0\right)\right)| \mathcal{D}\right], t \in(0,1)$. By taylor expansions, 
$$f(1)=f(0)+f^{\prime}(0)+f^{\prime \prime}(\tilde{t}) / 2, \quad \text { for some } \tilde{t} \in(0,1).$$
Hence, it suffices to show that 
\begin{equation}\label{prime}
f^{\prime}(0) = o_p(n^{-1/2})
\end{equation}
and 
\begin{equation}\label{pprime}
f^{\prime \prime}(\tilde{t}) = o_p(n^{-1/2}).
\end{equation}

To prove (\ref{prime}), we need to verify the Neyman orthogonality \citep{Neyman1959optimal,Neyman1979ca}, which is crucial for AIPW method. 
Denote $\mathcal V$ as the set of candidate value for the nuisance function $\omega$, and $\mathcal V_n$ as the set for estimators of $\omega$ satisfying Assumption A2 for any $\omega\in \mathcal V_n$ and $\mathcal V_n \subset \mathcal V$. Then 
$\widehat{\omega} \in \mathcal V_n$.
For any $\omega \in \mathcal{V}$, the Gateaux derivative in the direction $\omega-\omega_0 = (m-m_0, b-b_0)$ is
$$
\begin{aligned}
\partial_\omega \mathbb{E}\left[\Psi\left(W ; \omega_0\right)\right]\left[\omega-\omega_0\right]=& \mathbb{E}\left[m(1, X)-m_0(1, X)\right] - \mathbb{E}\left[m(0, X)-m_0(0, X)\right] \\
& - \mathbb{E}\left[\frac{A(m(1,X)- m_0(1,X))} {b_0(1,X)} \right] + 
\mathbb{E}\left[\frac{(1-A)(m(0,X)- m_0(0,X))} {b_0(0,X)} \right] \\
& -\mathbb{E}\left[\frac{A(Y- m_0(1,X)) (b(1,X)-b_0(1,X))} {b_0^2(1,X)} \right]  \\ &- \mathbb{E}\left[\frac{(1-A)(Y- m_0(0,X)) (b(0,X)-b_0(0,X))} {b_0^2(0,X)} \right].
\end{aligned}
$$
Since the Neyman orthogonality holds, that is, $\partial_\omega \mathbb{E}\left[\Psi\left(W ; \omega_0\right)\right]\left[\omega-\omega_0\right]=0$. So $f^{\prime}(0) = 0$.

To prove (\ref{pprime}), 
for any $t \in (0,1)$,
$$
\begin{aligned}
\partial^2 f(t)= & \mathbb{E}\left[\frac{A\left(m(1, X)-m_0(1, X)\right)\left(b(1,X)-b_0(1,X)\right)}{\left(b_0(1,X)+t\left(b(1,X)-b_0(1,X)\right)\right)^2}\right] \\
& +\mathbb{E}\left[\frac{(1-A)\left(m(0, X)-m_0(0, X)\right)\left(b(0,X)-b_0(0,X)\right)}{\left(b_0(0,X)-t\left(b(0,X)-b_0(0,X)\right)\right)^2}\right] \\
& +\mathbb{E}\left[\frac{\left(m(1, X)-m_0(1, X)\right)\left(b(1,X)-b_0(1,X)\right)}{\left(b_0(1,X)+t\left(b(1,X)-b_0(1,X)\right)\right)^2}\right] \\
& +2 \mathbb{E}\left[\frac{A\left(Y-m_0(1, X)-t\left(m(1, X)-m_0(1, X)\right)\right)\left(b(1,X)-b_0(1,X)\right)^2}{\left(b_0(1,X)+t\left(b(1,X)-b_0(1,X)\right)\right)^3}\right] \\
& +\mathbb{E}\left[\frac{\left(g(0, X)-g_0(0, X)\right)\left(b(0,X)-b_0(0,X)\right)}{\left(b_0(0,X)+t\left(b(0,X)-b_0(0,X)\right)\right)^2}\right] \\
& -2 \mathbb{E}\left[\frac{(1-A)\left(Y-g_0(0, X)-t\left(g(0, X)-g_0(0, X)\right)\right)\left(b(0,X)-b_0(0,X)\right)^2}{\left(b_0(0,X)+t\left(b(0,X)-b_0(0,X)\right)\right)^3}\right],
\end{aligned}
$$
\begin{comment}
$$
\begin{aligned}
\partial^2 f(t)= & 2\mathbb{E}\left[\frac{A\left(m(1, X)-m_0(1, X)\right)\left(b(1,X)-b_0(1,X)\right)}{\left(b_0(X)+t\left(b(X)-b_0(X)\right)\right)^2}\right] \\
& +2 \mathbb{E}\left[\frac{A\left(Y-m_0(1, X)-t\left(m(1, X)-m_0(1, X)\right)\right)\left(b(1,X)-b_0(1,X)\right)^2}{\left(b_0(1,X)+t\left(b(1,X)-b_0(1,X)\right)\right)^3}\right], \\
\end{aligned}
$$  
\end{comment}
Given $\mathbb{E}(Y-m_0(A,X)|A,X) = 0$, we have
$$\mid \partial^2 f(t)\mid  \leq C\sqrt{\mathbb{E}|m-m_0|^2}
\sqrt{\mathbb{E}|b-b_0|^2} = O_p(n^{-(l_1+l_2)}) = o_p(n^{-1/2}),$$
where $C$ is a constant. So (\ref{pprime}) holds.

Combing (\ref{prime}) and (\ref{pprime}), and based on Lemma \ref{uncondition} , we have proved that
$$\sup_{\omega \in \mathcal V_n, t\in (0,1)} \mid\partial^2 \Psi\left(W ; \omega_0+t\left(\omega-\omega_0\right)\right)\mid = o_p(n^{-1/2}). $$
Thus $\tilde\mu= \mu + o_p(n^{-1/2})$.

As for the asymptotic property of $\tilde\sigma^2$. Denote $\tilde\sigma_0 = \sqrt{1+\tilde\mu^2/\tilde\sigma^2}, \tilde\kappa_n = \frac{\sqrt{n}\tilde\mu}{\tilde\sigma}$. Then we have
$$ \tilde{\sigma}_0^2-\sigma_0^2=\frac{\tilde{\mu}^2}{\tilde{\sigma}^2}-\frac{\mu^2}{\sigma^2}=\frac{\tilde{\mu}^2 \sigma^2-\mu^2 \tilde\sigma^2}{\tilde{\sigma}^2  \sigma^2} =\frac{\tilde{\mu}^2\left(\sigma^2-\tilde{\sigma}^2\right)+\tilde{\sigma}^2\left(\tilde{\mu}^2-\mu^2\right)}{\tilde{\sigma}^2 \sigma^2}  $$
and
$$
\tilde{\kappa}_n-\kappa =\frac{\sqrt{n} \tilde{\mu}}{\tilde\sigma}-\frac{\sqrt{n} \mu}{\sigma} =\frac{\sqrt{n} \tilde{\mu} \sigma-\sqrt{n} \mu \tilde\sigma}{\tilde{\sigma} \sigma} 
=\frac{\sqrt{n}[(\tilde\mu-\mu) \sigma+\mu(\sigma-\tilde{\sigma})]}{\tilde{\sigma} \sigma}.
$$
Since
$$\tilde\mu = \mu + o_p(n^{-1/2}),$$
it suffices to prove
\begin{equation}\label{sigma}
\sigma^2 = O_p(1)
\end{equation}
and 
\begin{equation}\label{tildesigma}
\tilde\sigma - \sigma = o_p(n^{-1/2}) \text{  i.e.  }  \tilde\sigma^2 - \sigma^2 = (\tilde\sigma - \sigma)(\tilde\sigma + \sigma) = o_p(n^{-1/2}).
\end{equation}

As for Equation (\ref{sigma}),
$$
\begin{aligned}
\sigma^2 = & \operatorname{Var}\left(m(1,X)-m(0,X)+\frac{\mathbb{I}(A=1)}{b(1,X)} [Y-m(A,X)]-\frac{\mathbb{I}(A=0)}{b(0,X)} [Y-m(A,X)]\right) \\
= & \mathbb{E}\left[\operatorname{Var}(m(1,X)-m(0,X)+\frac{\mathbb{I}(A=1)}{b(1,X)} [Y-m(A,X)]-\frac{\mathbb{I}(A=0)}{b(0,X)} [Y-m(A,X)]\mid X, A)\right]\\
&+ \operatorname{Var}\left[\mathbb{E}(m(1,X)-m(0,X)+\frac{\mathbb{I}(A=1)}{b(1,X)} [Y-m(A,X)]-\frac{\mathbb{I}(A=0)}{b(0,X)} [Y-m(A,X)] \mid X, A)\right] \\
= & \mathbb{E}\left\{\left[\left(\frac{\mathbb{I}(A=1)}{b(1,X)}\right)^2+\left(\frac{\mathbb{I}(A=0)}{b(0,X)}\right)^2\right] \operatorname{Var}(Y)\right\}+\operatorname{Var}\left(m(1, X)-m(1, X)\right) \\
= & \operatorname{Var}(Y)\mathbb{E}\left[\mathbb{E}\left( \left(\frac{\mathbb{I}(A=1)}{b(1,X)}\right)^2+\left(\frac{\mathbb{I}(A=0)}{b(0,X)}\right)^2 \mid X\right)\right]+\operatorname{Var}\left(m(1, X)-m(0, X)\right) \\
= & \operatorname{Var}(Y)\mathbb{E}\left[ \frac1{b(1,X)} + \frac1{b(0,X)}\right]+\operatorname{Var}\left(m(1, X)-m(0, X)\right)\\
= & O(1).
\end{aligned}
$$

To prove Equation (\ref{tildesigma}), we only need to prove that $\tilde{\sigma}_1^2 = \sigma^2 + o_p(n^{-1/2})$.  Note that
$$
\begin{aligned}
& \tilde{\sigma}^2_1=\mathbb{E}(\widehat\mu_i^2 |  \mathcal{D})- \tilde\mu^2, \\
& \sigma^2=\mathbb{E}\left[\psi\left(W; w_0\right)\right]^2-\mu^2.
\end{aligned}
$$
\begin{comment}
Denote
{\small$$
\begin{aligned}
g(t):& =m_0(1, X)+t\left(m(1, X)-m_0(1, X)\right)+\frac{\mathbb{I}(A=1)}{b_0(1,X)+t\left(\left(b(1,X)-b_0(1,X)\right)\right.}\left[Y-m_0(1, X)-t\left(m(1, X)-m_0(1, X)\right)\right] \\
& -m_0(0, X)-t\left(m(0, X)-m_0(0, X)\right)-\frac{\mathbb{I}(A=0)}{b(0,X)-t\left(b(0,X)-b_0(0,X)\right)}\left[Y-m_0(0, X)-t\left(m(0, X)-m_0(0, X)\right)\right],
\end{aligned}
$$}  
\end{comment}
Denote $g(t) = \psi\left(W ; \omega_0+t\left(\omega-\omega_0\right)\right)$
and $h(t) = \mathbb{E}(g^2(t))$ with nuisance functions in $g(t)$ estimated via external data $\mathcal{D}$. By taylor expansions, 
$$h(1)=h(0)+h^{\prime}(0)+h^{\prime \prime}(\tilde{t}) / 2, \quad \text { for some } \tilde{t} \in(0,1).$$
Hence, it suffices to show that 
\begin{equation*}
h^{\prime}(0) = \mathbb{E}\left(2g(0)g^\prime(0)\right) = o_p(n^{-1/2})
\end{equation*}
and 
\begin{equation*}
h^{\prime \prime}(\tilde{t}) = \mathbb{E}\left(2g^\prime(t)^2 + 2g(t)g^{\prime\prime}(t) \right) = o_p(n^{-1/2}).
\end{equation*}
Note that
$$
\begin{aligned}
g^{\prime}(t)= & m(1, X)-m_0(1, X)-\frac{\mathbb{I}(A=1)\left(b(1,X)-b_0(1,X)\right) \left[Y-m_0(1, X)-t\left(m(1, X)-m_0(1, X)\right)\right]}{(b_0(1,X)+t\left(b(1,X)-b_0(1,X))\right)^2} \\
& -\frac{\mathbb{I}(A=1)\left[m(1, X)-m_0(1, X)\right]}{b_0(1,X)+t\left(b(1,X)-b_0(1,X)\right)}  -\left[m(0, X)-m_0(0, X)\right] \\
& +\frac{\mathbb{I}(A=0)\left(b(0,X)-b_0(0,X)\right) \left[Y-m_0(0, X)-t\left(m(0, X)-m_0(0, X)\right)\right]}{\left(b_0(0,X)-t\left(b(0,X)-b_0(0,X)\right)\right)^2}\\
& +\frac{\mathbb{I}(A=0)\left[m(0, X)-m_0(0, X)\right]}{b(0,X)-t\left(b(0,X)-b_0(0,X)\right)},
\end{aligned}
$$
and
$$
\begin{aligned}
g^{\prime \prime}(t)= & 2 \frac{\mathbb{I}(A=1)\left(b(1,X)-b_0(1,X)\right)^2 \left[Y-m_0(1, X)-t\left(m(1, X)-m_0(1, X)\right)\right]}{\left(b_0(1,X)+t\left(b(1,X)-b_0(1,X)\right)\right)^3} \\
& +2\frac{\mathbb{I}(A=1)\left(b(1,X)-b_0(1,X)\right)\left(m(1, X)-m_0(1, X)\right)}{\left(b_0(1,X)+t\left(b(1,X)-b_0(1,X)\right)\right)^2}\\
& -2 \frac{\mathbb{I}(A=0)\left(b(0,X)-b_0(0,X)\right)^2 \left[Y-m_0(0, X)-t\left(m(0, X)-m_0(0, X)\right)\right]}{\left(b_0(0,X)-t\left(b(0,X)-b_0(0,X)\right)\right)^3} \\
& +2\frac{\mathbb{I}(A=0)\left(b(0,X)-b_0(0,X)\right)\left(m(0, X)-m_0(0, X)\right)}{\left(b_0(0,X)-t\left(b(0,X)-b_0(0,X)\right)\right)^2}.
\end{aligned}
$$
Since $g^\prime(t) = 0$ and $\mathbb{E}({g(t)g^{\prime\prime}(t)}) = o_p(n^{-1/2})$, 
we have $h^\prime(t) = 0$ and $\mathbb{E}(h^{\prime\prime}(t)) = o_p(n^{-1/2})$ for $t \in (0,1)$. Thus we complete the proof for $\tilde\sigma^2$. \\

\noindent{\it Step II}. Denote the rejection region as $R_\alpha = (-\infty, z_{\alpha/2}) \cup (z_{1-\alpha/2}, +\infty)$. We prove that the pseudo outcome-based $T(\bar\pi^*)$ is valid and consistent under the null and alternative hypothesis, respectively.
On one hand, when $\mu<0$, $\lim_{n\rightarrow\infty}\tilde\kappa_n = -\infty$, making 
$$\lim_{n\rightarrow\infty} \mathbb{P}(T({\bar\pi}^*) \in R_\alpha) = \lim_{n\rightarrow\infty} 1 - \Phi(-\tilde\kappa_n + \frac {z_{1-\alpha/2}}{\tilde\sigma_0}) + e^{\frac{2\tilde\kappa_n z_{1-\alpha/2}}{\tilde\sigma_0}} \Phi(-\tilde\kappa_n - \frac {z_{1-\alpha/2}}{\tilde\sigma_0}) = 0.$$ 
When $\mu=0$, $\lim_{n\rightarrow\infty}\tilde\kappa_n = 0$, and we have  
$$\lim_{n\rightarrow\infty} \mathbb{P}(T({\bar\pi}^*) \in R_\alpha)) = \alpha. $$
So to sum up, when $\mu\leq0$, $\lim_{n\rightarrow\infty}\mathbb{P}(T({\bar\pi}^*\in R_\alpha) \leq \alpha$.

On the other hand, when $\mu>0$, $\lim_{n\rightarrow\infty} \tilde\kappa_n = \infty$. Therefore $$\lim_{n\rightarrow\infty} \mathbb{P}(T({\bar\pi}^*) \in R_\alpha) = \lim_{n\rightarrow\infty} 1 - \Phi(-\tilde\kappa_n + \frac {z_{1-\alpha/2}}{\tilde\sigma_0}) + e^{\frac{2\tilde\kappa_n z_{1-\alpha/2}}{\tilde\sigma_0}} \Phi(-\tilde\kappa_n - \frac {z_{1-\alpha/2}}{\tilde\sigma_0}) = 1.$$\\

\noindent{\it Step III}. In this step, we prove the efficiency of the permutation-based procedure.
Denote $\pi(u)=\frac{1}{B} \sum_b I\left\{p_{b} \leqslant u\right\}$. 

Under the null hypothesis,
$$
\lim_{n\rightarrow\infty}P(Q(\gamma) \leqslant \alpha) =\lim_{n\rightarrow\infty} E\left[I_{\{Q(\gamma) \leq \alpha\}}\right]=\lim_{n\rightarrow\infty} E\left[I_{\{\pi(\alpha \gamma) \geqslant \gamma\}}\right] \leqslant \frac{1}{\gamma} E(\pi(\alpha \gamma))\leq \alpha.
$$
Under the fixed alternative hypotheses, 
    $$
\lim_{n\rightarrow\infty}P(Q(\gamma) \leqslant \alpha) =\lim_{n\rightarrow\infty} E\left[I_{\{Q(\gamma) \leq \alpha\}}\right]=\lim_{n\rightarrow\infty} E\left[I_{\{\pi(\alpha \gamma) \geqslant \gamma\}}\right] = 1.
$$
Therefore, based on Steps I, II and III, the proof of Theorem 1 is completed.

\subsection{Proof of Theorem 2} \label{subsec:theoremDRL}
Denote $\bar\mu^{a} = \sum_{t=1}^T\mathbb{E}^a(Y_t) = \mathbb{E}[\sum_{k=1}^T {\omega}_k^a(X_k,A_k)(Y_k-{V}_k^a(X_k))+{\omega}_{k-1}^a(X_{k-1},A_{k-1}) {V}_k^a(X_k)]$, $\bar\sigma^{2,a} = \text{Var}(\sum_{k=1}^T {\omega}_k^a(X_k,A_k)(Y_k-{V}_k^a(X_k)+{\omega}_{k-1}^a(X_{k-1},A_{k-1}) {V}_k^a(X_k))$ for $a = 0,1$, where $\mathbb{E}^a$ denotes that the rewards are obtained when the policy $a$ is applied the whole day. Let
\begin{equation}\label{mu_a}
    \widehat{\mu}^a 
    = \widehat{V}_1^a(X_{1})
    + \sum_{k=1}^T\widehat{\omega}_{k}^{a}(X_{k}, A_{k}) \big[ Y_k + \widehat{V}_{k+1}^{a}(X_{k+1}) - \widehat{V}_k^{a}(X_{k}) \big],
\end{equation}
for $a=0,1$, where estimators for  ${V}_k^a(X_k)$s and ${\omega}_k^a(X_k,A_k)$s, i.e., $\widehat{V}_k^a(X_k)$s and $\widehat{\omega}_k^a(X_k,A_k)$s, are estimated via external dataset.
To prove this theorem, we only need to verify that
$\widehat{\mu}^a$ satisfies
\begin{equation}\label{drlmu}
    \mathbb{E}(\widehat{\mu}^a) = \bar\mu^a + o_p(n^{-1/2}),
\end{equation}
and
\begin{equation}\label{drlsigma}
    \text{Var}(\widehat{\mu}^a) = \bar\sigma^{2,a} + o_p(n^{-1/2})
\end{equation}
for $a=0,1$.
Then following the steps II and III in the proof of Theorem 1, we can proof Theorem 2.

Note that Equation (\ref{mu_a}) equals to 
$$ \widehat\mu^a = \sum_{k=1}^T \widehat{\omega}_k^a(X_{k},A_{k})(Y_k-\widehat{V}_k^a(X_k))+\widehat{\omega}_{k-1}^a(X_{k-1},A_{k-1}) \widehat{V}_k^a(X_k).$$
Then
{\allowdisplaybreaks
\begin{align*}
\mathbb{E}(\widehat\mu^a) - \bar\mu^a  & = \mathbb{E}[\sum_{k=1}^T \widehat{\omega}_k^a(X_k,A_k)(Y_k-\widehat{V}_k^a(X_k))+\widehat{\omega}_{k-1}^a(X_{k-1},A_{k-1}) \widehat{V}_k^a(X_k)] \\
&- \mathbb{E}[\sum_{k=1}^T {\omega}_k^a(X_k,A_k)(Y_k-{V}_k^a(X_k))+{\omega}_{k-1}^a(X_{k-1}, A_{k-1}) {V}_k^a(X_k)] \\
& = \mathbb{E}\left[\sum_{k=1}^T\left(\widehat{\omega}_{k}^a(X_{k}, A_{k})-\omega_k^a(X_k,A_k)\right)\left(-\widehat{V}_k^{a}(X_k)+V_k^a(X_k)\right)\right.\\
& + \left.\left(\widehat{\omega}_{k-1}^{a}(X_{k-1},A_{k-1})-\omega_{k-1}^a(X_{k-1},A_{k-1})\right)\left(\widehat{V}_k^{a}(X_k)-V_k^a(X_k)\right) \right] \\
& + \mathbb{E}\left[\sum_{k=1}^T \omega_k^a(X_k,A_k)\left(-\widehat{V}_k^{a}(X_k)+V_k^a(X_k)\right)+\omega_{k-1}^a(X_{k-1},A_{k-1})\left(\widehat{V}_{k}^a(X_{k})-V_k^a(X_k)\right) \right] \\
& + \mathbb{E}\left[\sum_{k=1}^T\left(\widehat{\omega}_{k}^a(X_{k}, A_{k})-\omega_k^a(X_k,A_k)\right)\left(Y_k-V_k^a(X_k)+V_{k+1}^a(X_{k+1})\right) \right] \\
& = \mathbb{E}\left[\sum_{k=1}^T\left(\widehat{\omega}_{k}^a(X_{k}, A_{k})-\omega_k^a(X_k,A_k)\right)\left(-\widehat{V}_{k}^a(X_{k})+V_k^a(X_k)\right) \right.\\
&\left.+\left(\widehat{\omega}_{k-1}^{a}(X_{k-1},A_{k-1})-\omega_{k-1}^a(X_{k-1},A_{k-1})\right)\left(-\widehat{V}_{k}^a(X_{k})+V_k^a(X_k)\right) \right] \\
& = \sum_{k=1}^T \mathrm{O}\left(\left\|\widehat{\omega}_{k}^a(X_{k}, A_{k})-\omega_k^a(X_k,A_k)\right\|_2\left\|\widehat{V}_{k}^a(X_{k})-V_k^a(X_k)\right\|_2\right) \\
& = \mathrm{o}_p(n^{-1 / 2}).
\end{align*}
}
Thus Equation (\ref{mu_a}) is proved.

$$
\begin{aligned}
    \text{Var}(\widehat\mu^a) - \bar\sigma^{2,a} & = \mathbb{E}[(\widehat\mu^a)^2]+[\mathbb{E}(\widehat\mu^a)]^2 \\
    &- \mathbb{E}\left[\Big(\sum_{k=1}^T {\omega}_k^a(X_k,A_k)(Y_k-{V}_k^a(X_k))+{\omega}_{k-1}^a(X_{k-1},A_{k-1}) {V}_k^a(X_k)\Big)^2\right]  - (\bar\mu^a)^2\\
    & = \mathbb{E}\left\{\left[\widehat\mu^a - \Big(\sum_{k=1}^T {\omega}_k^a(X_k,A_k)(Y_k-{V}_k^a(X_k))+{\omega}_{k-1}^a(X_{k-1},A_{k-1}) {V}_k^a(X_k)\Big) \right] \right. \\
    & \left. \times \left[\widehat\mu^a + \Big(\sum_{k=1}^T {\omega}_k^a(X_k,A_k)(Y_k-{V}_k^a(X_k))+{\omega}_{k-1}^a(X_{k-1},A_{k-1}) {V}_k^a(X_k)\Big) \right]\right\} \\
    & + [\mathbb{E}(\widehat\mu^a) - \bar\mu^a][\mathbb{E}(\widehat\mu^a) + \bar\mu^a] \\
    & \leq \mathbb{E}\bigg\{\bigg[\widehat\mu^a - \Big(\sum_{k=1}^T {\omega}_k^a(X_k,A_k)(Y_k-{V}_k^a(X_k))+{\omega}_{k-1}^a(X_{k-1},A_{k-1}) {V}_k^a(X_k)\Big) \bigg]^2\bigg\}^{1/2}  \\
    &\times \mathbb{E}\bigg\{\bigg[\widehat\mu^a + \Big(\sum_{k=1}^T {\omega}_k^a(X_k,A_k)(Y_k-{V}_k^a(X_k))+{\omega}_{k-1}^a(X_{k-1},A_{k-1}) {V}_k^a(X_k)\Big) \bigg]^2\bigg\}^{1/2} \\
    & + [\mathbb{E}(\widehat\mu^a) - \bar\mu^a][\mathbb{E}(\widehat\mu^a) + \bar\mu^a] \\ 
    & = C_1\sqrt{\mathbb{E}(A_1 + A_2 + A_3)^2} + C_2[\mathbb{E}(\widehat\mu^a) - \bar\mu^a]
\end{aligned}
$$
with $C_1$ and $C_2$ being finite constants determined by the boundedness of rewards, and 
$$
\begin{aligned}
A_1 & = \left[\sum_{k=1}^T\left(\widehat{\omega}_{k}^a(X_{k}, A_{k})-\omega_k^a(X_k,A_k)\right)\left(-\widehat{V}_{k}^a(X_{k})+V_k^a(X_k)\right) \right. \\ & \left.+\left(\widehat{\omega}_{k-1}^{a}(X_{k-1},A_{k-1})-\omega_{k-1}^a(X_{k-1},A_{k-1})\right)\left(\widehat{V}_{k}^a(X_{k})-V_k^a(X_k)\right) \right],\\
A_2 & = \left[\sum_{k=1}^T \omega_k^a(X_k,A_k)\left(-\widehat{V}_{k}^a(X_{k})+V_k^a(X_k)\right)+\omega_{k-1}^a(X_{k-1},A_{k-1})\left(\widehat{V}_{k}^a(X_{k})-V_k^a(X_k)\right) \right], \\
A_3 &= \left[\sum_{k=1}^T\left(\widehat{\omega}_{k}^a(X_{k}, A_{k})-\omega_k^a(X_k,A_k)\right)\left(Y_k-V_k^a(X_k)+V_{k+1}^a(X_{k+1})\right) \right].
\end{aligned}
$$
Therefore,
$$
\begin{aligned}
  \text{Var}(\widehat\mu^a) - \bar\sigma^{2,a} & \leq \sqrt{\mathbb{E}(A_1^2 + A_2^2 + A_3^2 + 2A_1A_2 + 2A_1A_3 + 2A_2A_3)} + o(n^{-1/2}) = o(n^{-1/2}).  
\end{aligned}
$$
The proof is completed.

\section{Simulations} \label{sec:numerical}

In this section, we conduct numerical simulations to compare the proposed test with existing tests. We first consider completely randomized studies in Section \ref{subsec:random}. We next investigate confounded observational studies in Section \ref{subsec:confounded}. Finally, dynamic settings are considered in Section \ref{subsec:sequentialsimulation}.

\subsection{Completely randomized study} \label{subsec:random}
In this subsection, we conduct 
simulations to investigate the finite sample performance of the proposed methods. 
The covariates-treatment-outcome triplet is generated as follows:
\begin{itemize}
    \item \textbf{Covariates}: The baseline covariates, $X=(X_1,X_2)$, are two-dimensional and follow a mean-zero Gaussian distribution with identity covariance matrix. 
    \item \textbf{Treatment}: A completely randomized study is considered where the treatment is generated independently of any baseline covariates. Specifically, $A$ follows a Bernoulli distribution with success probability $p_a\in \{0.3,0.5\}$. 
    \item \textbf{Outcome}: The outcome $Y$ is generated according to $Y=(X_1-X_2 + 2)/2+A\tau(X)+\varepsilon$, where $\tau(X)$ corresponds to the conditional ATE (CATE), quantifying the difference between the two potential outcomes given $X$. $\varepsilon$ is a Gaussian noise term with a mean of zero and standard deviation $\sigma_0$, which takes values from $\{0.5,1,3\}$. 
\end{itemize}
We consider two null hypotheses $\mathcal{H}_0^{(1)}$, $\mathcal{H}_0^{(2)}$ and three alternative hypotheses $\mathcal{H}_1^{(1)}$, $\mathcal{H}_1^{(2)}$ and $\mathcal{H}_1^{(3)}$, defined by the form of the CATE $\tau(X)$, as follows:
\begin{itemize}
    \item $\mathcal{H}_0^{(1)}: \tau(X)\equiv 0$,
    \item $\mathcal{H}_0^{(2)}: \tau(X)=0.2^{\mathbb{I}(\sigma_0=0.5)} 0.3^{\mathbb{I}(\sigma_0=1)} \frac{\sqrt{\pi} }{16} (X_1+X_2)^3$,
    \item $\mathcal{H}_1^{(1)}: 0.2^{\mathbb{I}(\sigma_0=0.5)} 0.3^{\mathbb{I}(\sigma_0=1)}0.8\max(1, X_1+X_2)$,
    \item $\mathcal{H}_1^{(2)}: 
    0.2^{\mathbb{I}(\sigma_0=0.5)} 0.3^{\mathbb{I}(\sigma_0=1)}0.8|X_1+X_2|$,
    \item $\mathcal{H}_1^{(3)}: 0.2^{\mathbb{I}(\sigma_0=0.5)} 0.3^{\mathbb{I}(\sigma_0=1)}0.5(X_1+X_2)^2$.
\end{itemize}
\begin{comment}
\begin{eqnarray*}
&\mathcal{H}_0^{(1)}: \tau(X)\equiv 0, \,\, &\mathcal{H}_0^{(2)}: \tau(X)=0.2^{\mathbb{I}(\sigma=0.5)} 0.3^{\mathbb{I}(\sigma=1)} \frac{\sqrt{\pi} }{16} (X_1+X_2)^3, \\
    &\mathcal{H}_1^{(1)}: 
    0.2^{\mathbb{I}(\sigma=0.5)} 0.3^{\mathbb{I}(\sigma=1)}0.8|X_1+X_2|, 
    &\mathcal{H}_1^{(2)}: 0.2^{\mathbb{I}(\sigma=0.5)} 0.3^{\mathbb{I}(\sigma=1)}0.8\max(1, X_1+X_2), & 
\end{eqnarray*}
\end{comment}
Here, the first null hypothesis $\mathcal{H}_0^{(1)}$ corresponds to the sharp null indicating no treatment effect at all. In the second null hypothesis $\mathcal{H}_0^{(2)}$, the ATE equals zero, despite a nonzero CATE. In the three alternative hypotheses, CATE is almost surely positive, which in turn yields a positive ATE. Additionally, the scaling factor $0.2^{\mathbb{I}(\sigma_0=0.5)} 0.3^{\mathbb{I}(\sigma_0=1)}$ in the latter four hypotheses adjusts the magnitude of the CATE to ensure a consistent signal-to-noise ratio across various levels of the standard deviation, $\sigma_0$.

For each hypothesis, we fix the sample size $n$ to 300, and consider two different treatment assignment mechanisms with $p_a\in \{0.3,0.5\}$ as well as three levels of residual variance $\sigma_0^2\in \{0.5, 1, 3\}$. 
This configuration results in a total of 30 simulation settings.

For each setting, we conduct 500 repetitions to evaluate the empirical type-I error rates and the power of the following five tests: (i) P-TAB; (ii) TAB; (iii) DML; (iv) A kernel treatment effects-based test \citep[denoted by KTE]{muandet2021counterfactual} and (v) its variant based on cross U-statistics \citep[denoted by xKTE]{martinez2024efficient}. 
All tests are performed at a significance level of 0.05 and employ $5$-fold cross-fitting to construct the pseudo outcomes and to form the test statistics. 

\begin{figure}[ht]
    \centering
    \includegraphics[width = 12cm, height = 10cm]{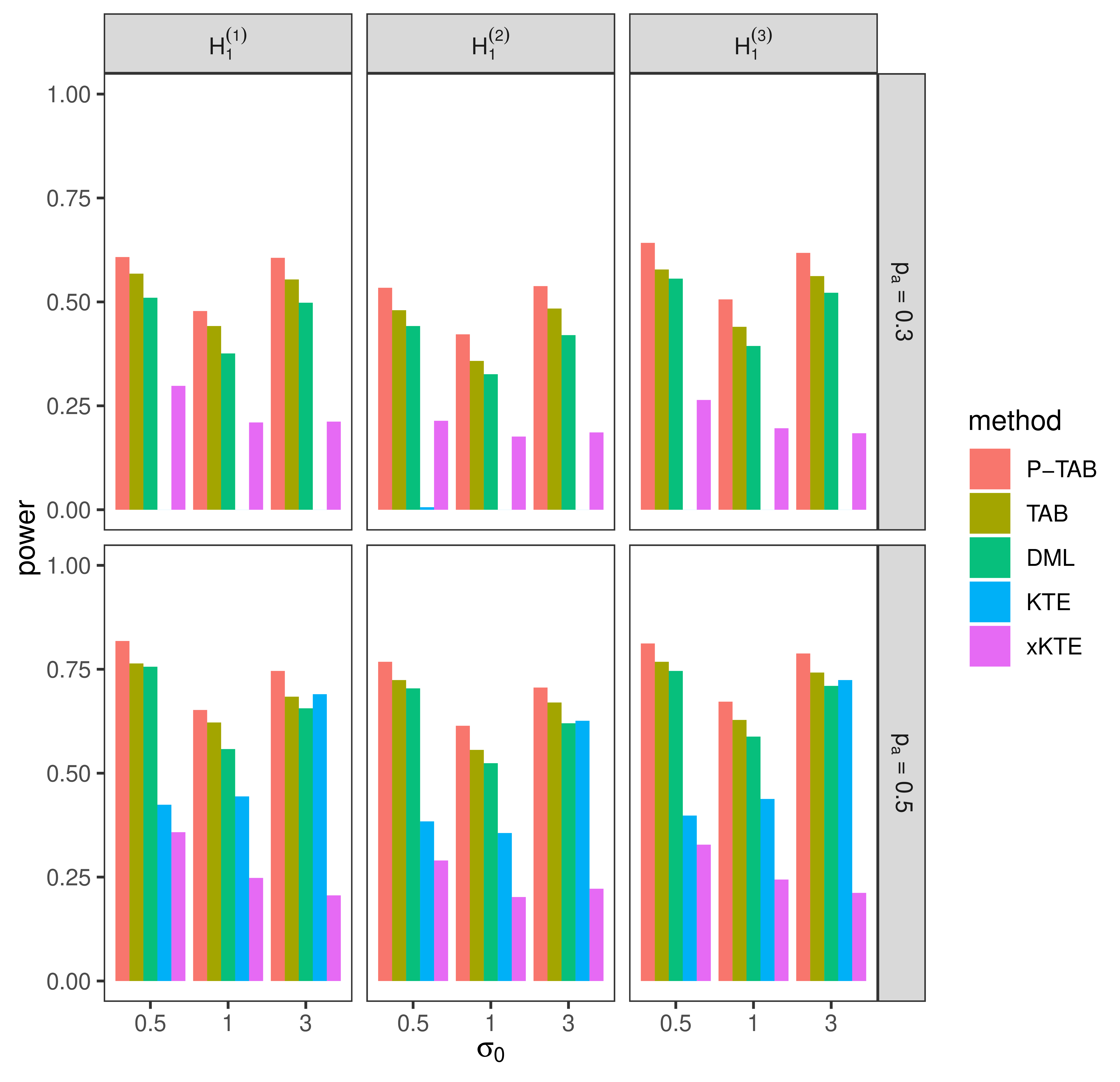}
    \caption{Powers of different methods in settings with unconfounded treatment assignment.}
    \label{pow-data1}
\end{figure}

The results under the null and alternative hypotheses are presented in Table \ref{typeI-data1} and Figure \ref{pow-data1}, respectively.  
It can be seen that P-TAB, TAB, DML and xKTE control the type-I error rates properly, while KTE suffers from inflated type-I error rates in settings where $p_a=0.5$. 
As for powers, it is evident that TAB achieves greater power than DML, which demonstrates the usefulness of the two-armed bandit-based test. Additionally, P-TAB demonstrates even greater power than TAB, showcasing the benefits of the permutation method within the TAB framework. These results align with our expectations. Finally, all the three methods outperform KTE and xKTE. Note that when $p_a = 0.3$, that is, when the sample is unbalanced between the two treatments, the powers of KTE is close to zero. 

\begin{remark}
    In addition to the aforementioned competitors, we also implemented the imputation- and IPW-based test. 
    Results (not reported here) 
    show that these tests fail to adequately control the type-I error. 
    Therefore, we have chosen not to present these results.
\end{remark}

\begin{table}[]
\centering
\caption{Type-I errors of different methods in settings with unconfounded treatment assignment.}
\label{typeI-data1}
\small
\begin{tabular}{llllllllllll}
\hline
\multicolumn{2}{c}{$\mathcal{H}_0$}                                             & \multicolumn{5}{c}{$\mathcal{H}_0^{(1)}$}                                                                                                                     & \multicolumn{5}{c}{$\mathcal{H}_0^{(2)}$}                                                                                                                     \\ \hline
\multicolumn{1}{c}{$p_a$}                    & \multicolumn{1}{c}{$\sigma_0$} & \multicolumn{1}{c}{P-TAB} & \multicolumn{1}{c}{TAB}   & \multicolumn{1}{c}{DML}   & \multicolumn{1}{c}{KTE}   & \multicolumn{1}{c}{xKTE}  & \multicolumn{1}{c}{P-TAB} & \multicolumn{1}{c}{TAB}   & \multicolumn{1}{c}{DML}   & \multicolumn{1}{c}{KTE}   & \multicolumn{1}{c}{xKTE}  \\ \hline
\multicolumn{1}{c}{\multirow{3}{*}{0.3}} & \multicolumn{1}{c}{0.5}   & \multicolumn{1}{c}{0.036} & \multicolumn{1}{c}{0.036} & \multicolumn{1}{c}{0.016} & \multicolumn{1}{c}{0.06}  & \multicolumn{1}{c}{0.056} & \multicolumn{1}{c}{0.048} & \multicolumn{1}{c}{0.038} & \multicolumn{1}{c}{0.034} & \multicolumn{1}{c}{0.05}  & \multicolumn{1}{c}{0.06}  \\
\multicolumn{1}{c}{}                     & \multicolumn{1}{c}{1}     & \multicolumn{1}{c}{0.056} & \multicolumn{1}{c}{0.05}  & \multicolumn{1}{c}{0.034} & \multicolumn{1}{c}{0.05}  & \multicolumn{1}{c}{0.076} & \multicolumn{1}{c}{0.032} & \multicolumn{1}{c}{0.034} & \multicolumn{1}{c}{0.016} & \multicolumn{1}{c}{0.058} & \multicolumn{1}{c}{0.058} \\
\multicolumn{1}{c}{}                     & \multicolumn{1}{c}{3}     & \multicolumn{1}{c}{0.048} & \multicolumn{1}{c}{0.038} & \multicolumn{1}{c}{0.028} & \multicolumn{1}{c}{0.048} & \multicolumn{1}{c}{0.052} & \multicolumn{1}{c}{0.046} & \multicolumn{1}{c}{0.034} & \multicolumn{1}{c}{0.024} & \multicolumn{1}{c}{0.046} & \multicolumn{1}{c}{0.056} \\ \hline
\multicolumn{1}{c}{\multirow{3}{*}{0.5}} & \multicolumn{1}{c}{0.5}   & \multicolumn{1}{c}{0.05}  & \multicolumn{1}{c}{0.05}  & \multicolumn{1}{c}{0.03}  & \multicolumn{1}{c}{0.088} & \multicolumn{1}{c}{0.054} & \multicolumn{1}{c}{0.042} & \multicolumn{1}{c}{0.046} & \multicolumn{1}{c}{0.024} & \multicolumn{1}{c}{0.084} & \multicolumn{1}{c}{0.048} \\
\multicolumn{1}{c}{}                     & \multicolumn{1}{c}{1}     & \multicolumn{1}{c}{0.042} & \multicolumn{1}{c}{0.036} & \multicolumn{1}{c}{0.022} & \multicolumn{1}{c}{0.064} & \multicolumn{1}{c}{0.052} & \multicolumn{1}{c}{0.04}  & \multicolumn{1}{c}{0.048} & \multicolumn{1}{c}{0.026} & \multicolumn{1}{c}{0.076} & \multicolumn{1}{c}{0.056} \\
\multicolumn{1}{c}{}                     & \multicolumn{1}{c}{3}     & \multicolumn{1}{c}{0.058} & \multicolumn{1}{c}{0.046} & \multicolumn{1}{c}{0.028} & \multicolumn{1}{c}{0.05}  & \multicolumn{1}{c}{0.062} & \multicolumn{1}{c}{0.048} & \multicolumn{1}{c}{0.038} & \multicolumn{1}{c}{0.026} & \multicolumn{1}{c}{0.068} & \multicolumn{1}{c}{0.056} \\ \hline
\end{tabular}
\end{table}

\subsection{Confounded observational study}\label{subsec:confounded}
We next consider settings where the treatment assignment is confounded by the baseline covariates, typically seen in observational studies. To systematically evaluate the proposed tests, we have designed five null hypotheses, denoted by $\{\mathcal{H}_0^{(j)}\}_{j=1}^5$, and five alternative hypotheses, denoted by $\{\mathcal{H}_1^{(j)}\}_{j=1}^5$. The covariates-treatment-outcome triplet is generated as follows: 
\begin{itemize}
    \item \textbf{Covariates}: 
    The covariate vector $X$ 
    is $d$ dimensional, with $d$ chosen from $\{3, 20, 50\}$. The first component $X_1$ is uniformly distributed between $(0,1)$, i.e., $X_1\sim U(0,1)$. The second and third variables are distributed according to a Bernoulli distribution with a success probability of $0.5$ under the first two alternative hypotheses, and they follow a uniform distribution $U(-2,2)$ under the remaining hypotheses. All other variables are independently sampled from a standard normal distribution.
    \item \textbf{Treatment}: The treatment $A$ follows a Bernoulli distribution with success probability $X_1$,i.e., $A\sim$~Bernoulli($X_1$). 
    \item \textbf{Outcome}: The outcome $Y$ satisfies $Y=m(0,X)+A\tau(X)+\varepsilon$, with two distributional types for $\varepsilon$ considered. 
    \begin{itemize}
        \item Normal distribution:

        \begin{itemize}
        \item The \textbf{baseline function $m(0,X)$} is fixed to \(X_2^2\) across all null hypotheses. Under the alternative hypotheses, it varies, equal to \(0.3\), \(0.024\), \(X_2\), \(X_2^2\), and \(X_2^2\) respectively.
        \item The \textbf{CATE} $\tau(X)$ equals $r \tau_0(X)$ where $\tau_0(X)$ is set to $X_2^2-4/3$, $X_1 X_2^2 X_3$, $2X_3\cos(\pi X_2/4)$, $2X_3\sin(\pi X_2/4)$, $2\sin(\pi X_2/4)$ under the null hypotheses and $0.48\mathbb{I}(X_1\le 0.5)$, $0.032$, $0.7 X_1 X_2^2$, $1.6 X_1 \cos(\pi X_2/4)$, $1.8X_1 \cos(\pi X_2/4)^2$ under the alternative hypotheses, and the scaling factor $r$  is fixed to $0.5^{\mathbb{I}(\sigma_0 = 0.5)} 0.8^{\mathbb{I}(\sigma_0 = 1)} 2.5^{\mathbb{I}(\sigma_0 = 3)}$ across all the hypotheses.
        \item The \textbf{residual} $\varepsilon$ 
        follows a normal distribution with standard deviation $\sigma_0$ chosen  from \(\{0.5, 1, 3\}\) under the null hypotheses and the last three alternative hypotheses. For the first two alternative hypotheses, \(\varepsilon\) incorporates an additional independent Bernoulli error term \(\text{Bernoulli}(\min (1, m(0,X)+A\tau(X)))-\min (1, m(0,X)+A\tau(X)) \). We also consider scenarios where $\epsilon$ follows a standard \(t\)-distribution with degrees of freedom chosen from \(\{3, 5, 10\}\) and list the detailed settings in the Appendix.
    \end{itemize}

        \item $t$ distribution:

        \begin{itemize}
        \item The \textbf{Baseline function $m(0,X)$} is fixed to \(X_2^2\) across all null hypotheses, and varies under the alternative hypotheses as \(0.3\), \(0.015\), \(X_2\), \(X_2^2\), and \(X_2^2\) respectively.
        \item The \textbf{CATE $\tau(X))$} $\tau(X)$ is set to $X_2^2-4/3$, $X_1 X_2^2 X_3$, $2X_3\cos(\pi X_2/4)$, $2X_3\sin(\pi X_2/4)$, $2\sin(\pi X_2/4)$ under the null hypotheses and $0.48\mathbb{I}(X_1\le 0.5)$, $0.1$, $0.8r X_1 X_2^2$, $2r X_1 \cos(\pi X_2/4)$, $2rX_1 \cos(\pi X_2/4)^2$ under the alternative hypotheses, with scaling factor $r = 2^{\mathbb{I}(df = 3)} 0.5^{\mathbb{I}(df = 10)}$ and df defined below.
        \item The \textbf{Residual $\varepsilon$} follows a $t$ distribution with degree of freedom (df) chosen  from \(\{3, 5, 10\}\) under the null hypotheses and the last three alternative hypotheses, \(\varepsilon\). For the first two alternative hypotheses, \(\varepsilon\) includes an additional independent Bernoulli error structured as \(\text{Bernoulli}(\min (1, m(0,X)+A\tau(X))) \). 
    \end{itemize}
    \end{itemize}

\end{itemize}
To summarize, we have 10 hypotheses (5 null and 5 alternative hypotheses), 3 choices for the dimensionality of covariates, and 6 residual distributions (3 normal and 3 $t$-distributions). This results in a total of 180 settings. 

Additionally, the dimension 
$d$ can be considerably larger than $3$, being either 20 or 50. However, only the first three variables are involved in the outcome regression or the propensity score function. 
To handle the high-dimensionality, we implement model-$X$ knockoffs \citep{candes2018panning} for variable selection in the continuous outcome regression function 
$m$. As for the propensity score function 
$b$ which involves binary outcome variables, we utilize the group method of data handling \citep{dag2019gmdh2} for variable selection. 
Both nuisance functions are then learned using the generalized boosted regression on the selected variables, to address the potential non-linearity in the data generating process.

\begin{figure}[h]
    \centering
    \includegraphics[height=8cm,width=0.7\textwidth]{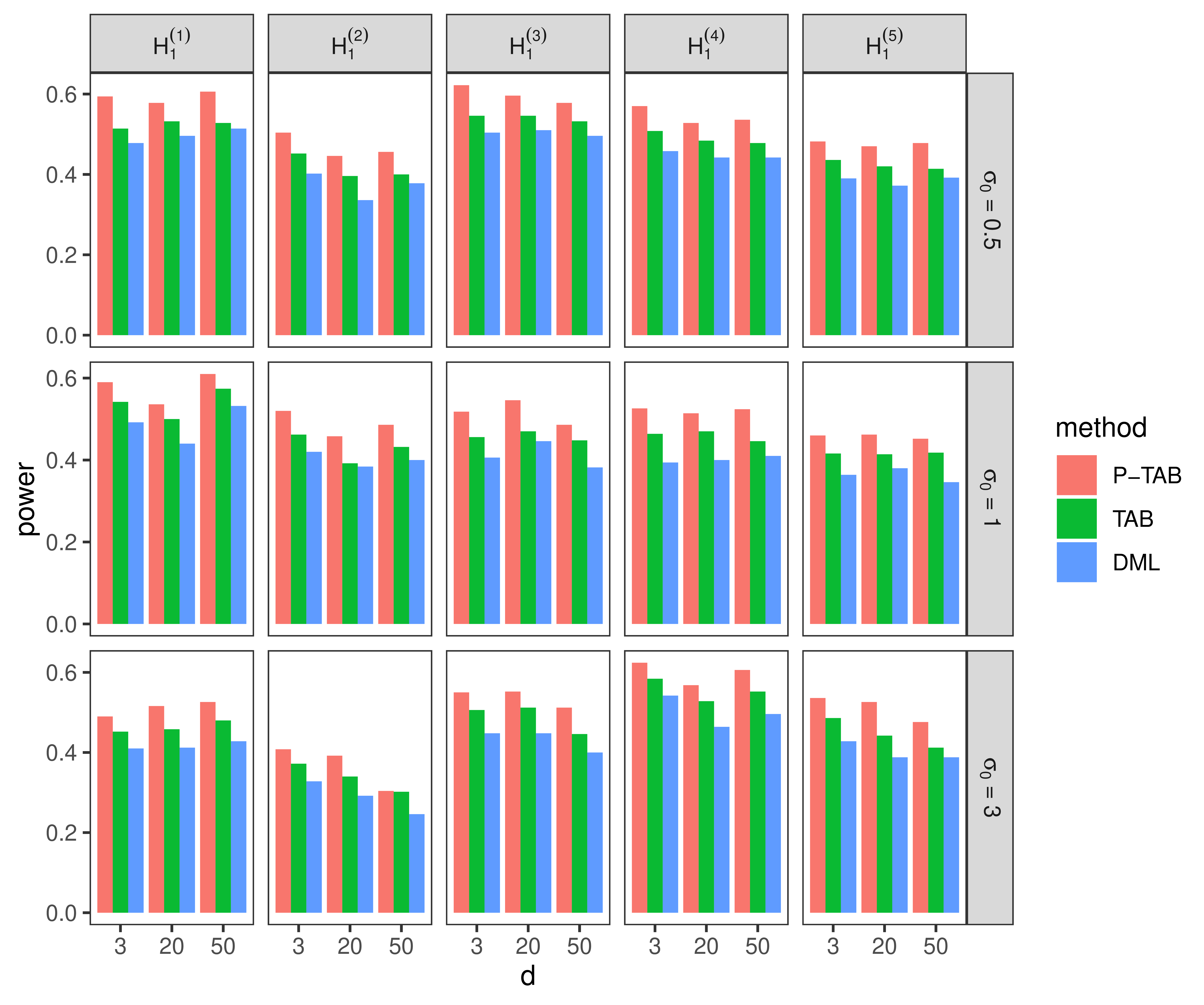}
    \caption{Powers of different methods with confounded treatment assignment and normally distributed error term.}
	\label{two-200-normal}
\end{figure}
\begin{figure}[t]
	\centering
		\includegraphics[height=8cm,width=0.7\textwidth]{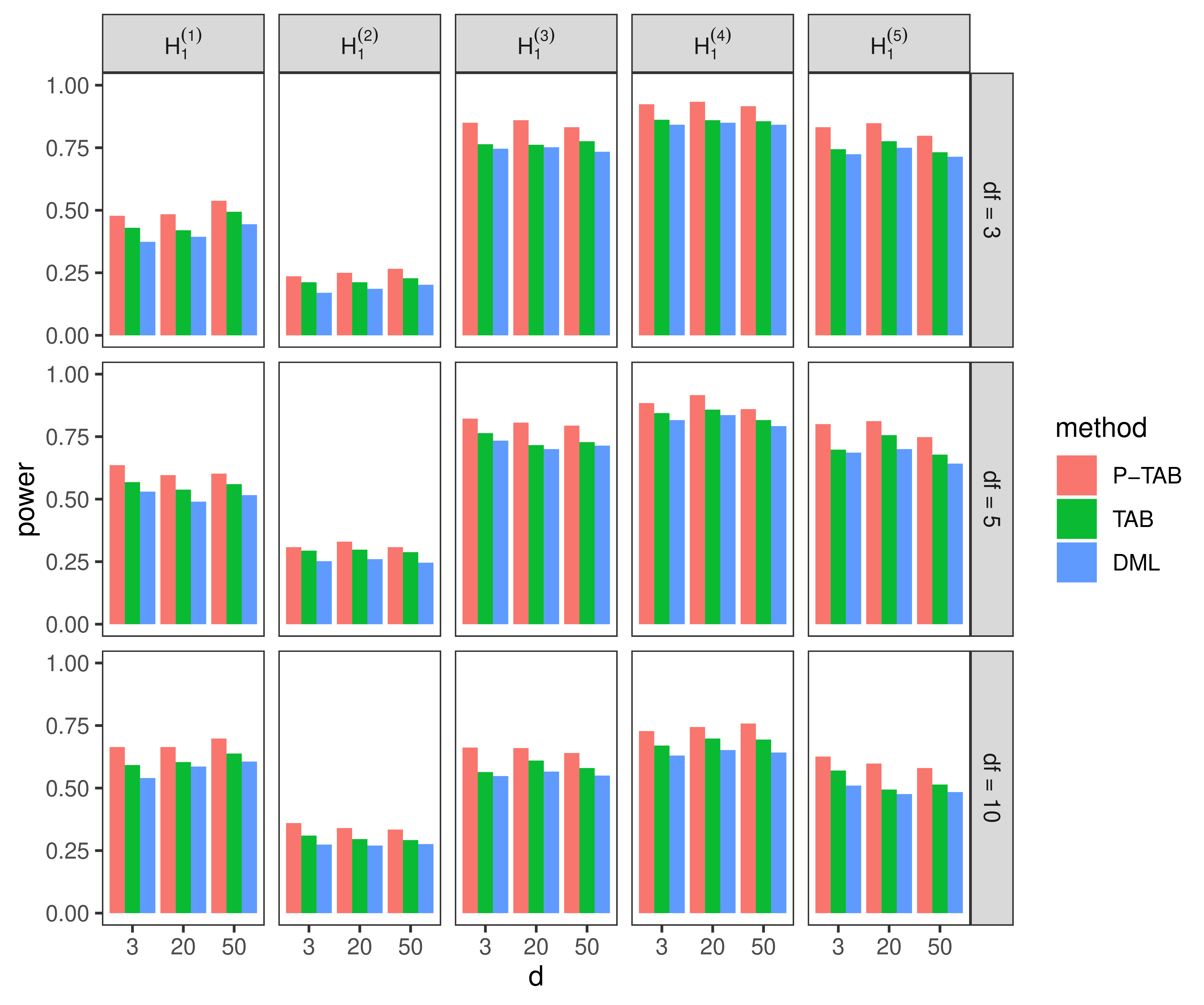}
 \caption{Powers of different methods with confounded treatment assignment and $t$ distributed error term.}
	\label{two-200-t}
\end{figure}

The type-I error rates are reported in Table \ref{typeI-data2}, and the powers are visualized in Figures \ref{two-200-normal} and \ref{two-200-t}. Our results are summarized as follows. First, notice that both KTE and xKTE suffer from inflated type-I error rates in these confounded settings. Therefore, their powers are meaningless and we did not report them in the figures. Second, P-TAB, TAB and DML effectively control the type-I error rates in almost all settings, regardless of whether the residuals follow normal or heavy-tailed distributions. Third, the proposed P-TAB, consistently achieves the highest statistical power across all scenarios. Finally, it is also worthwhile to emphasize that the variable selection procedure plays an important role in accurately estimating the nuisance functions $m$ and $b$ in high dimensions, which substantially aids in controlling the type-I error and enhancing power. 


\begin{table}[t]
\centering
\caption{Type-I errors of different methods in settings with confounded treatment assignment.}
\small
\label{typeI-data2}
\begin{tabular}{cccccccccccccc}
\hline
\multirow{2}{*}{$\mathcal{H}_0$} & \multirow{2}{*}{$d$}  & \multicolumn{6}{c}{normally distributed $\varepsilon$} & \multicolumn{6}{c}{t distributed $\varepsilon$}  \\ \cline{4-7} \cline{10-13}
                     &                     & $\sigma_0$  & P-TAB   & TAB    & DML    & KTE    & xKTE  & df & P-TAB  & TAB   & DML   & KTE   & xKTE  \\ \hline
\multirow{9}{*}{$\mathcal{H}_0^{1}$}   & \multirow{3}{*}{3}  & 0.5  & 0.02   & 0.034  & 0.016  & 0.072  & 0.81  & 3  & 0.046 & 0.05  & 0.04  & 0.078 & 0.764 \\
                     &                     & 1    & 0.036  & 0.048  & 0.028  & 0.086  & 0.79  & 5  & 0.038 & 0.052 & 0.024 & 0.058 & 0.79  \\
                     &                     & 3    & 0.066  & 0.066  & 0.042  & 0.074  & 0.608 & 10 & 0.058 & 0.064 & 0.042 & 0.07  & 0.86  \\
                     & \multirow{3}{*}{20} & 0.5  & 0.032  & 0.04   & 0.022  & 0.086  & 0.786 & 3  & 0.044 & 0.054 & 0.038 & 0.08  & 0.644 \\
                     &                     & 1    & 0.04   & 0.046  & 0.03   & 0.066  & 0.788 & 5  & 0.034 & 0.044 & 0.026 & 0.078 & 0.686 \\
                     &                     & 3    & 0.068  & 0.058  & 0.05   & 0.064  & 0.638 & 10 & 0.036 & 0.034 & 0.022 & 0.078 & 0.718 \\
                     & \multirow{3}{*}{50} & 0.5  & 0.03   & 0.038  & 0.022  & 0.116  & 0.432 & 3  & 0.05  & 0.054 & 0.032 & 0.09  & 0.424 \\
                     &                     & 1    & 0.06   & 0.054  & 0.034  & 0.088  & 0.436 & 5  & 0.058 & 0.048 & 0.038 & 0.092 & 0.504 \\
                     &                     & 3    & 0.046  & 0.058  & 0.038  & 0.108  & 0.326 & 10 & 0.06  & 0.074 & 0.058 & 0.106 & 0.536 \\ \hline
\multirow{9}{*}{$\mathcal{H}_0^{2}$}   & \multirow{3}{*}{3}  & 0.5  & 0.03   & 0.024  & 0.022  & 0.092  & 0.118 & 3  & 0.028 & 0.034 & 0.014 & 0.09  & 0.126 \\
                     &                     & 1    & 0.04   & 0.042  & 0.024  & 0.124  & 0.102 & 5  & 0.036 & 0.034 & 0.018 & 0.11  & 0.106 \\
                     &                     & 3    & 0.062  & 0.068  & 0.042  & 0.094  & 0.122 & 10 & 0.05  & 0.05  & 0.036 & 0.102 & 0.162 \\
                     & \multirow{3}{*}{20} & 0.5  & 0.03   & 0.032  & 0.024  & 0.102  & 0.106 & 3  & 0.058 & 0.044 & 0.042 & 0.082 & 0.106 \\
                     &                     & 1    & 0.026  & 0.018  & 0.014  & 0.12   & 0.102 & 5  & 0.056 & 0.046 & 0.04  & 0.104 & 0.086 \\
                     &                     & 3    & 0.068  & 0.052  & 0.042  & 0.128  & 0.138 & 10 & 0.042 & 0.046 & 0.032 & 0.1   & 0.088 \\
                     & \multirow{3}{*}{50} & 0.5  & 0.04   & 0.036  & 0.036  & 0.11   & 0.048 & 3  & 0.048 & 0.05  & 0.038 & 0.114 & 0.06  \\
                     &                     & 1    & 0.044  & 0.062  & 0.034  & 0.114  & 0.048 & 5  & 0.06  & 0.066 & 0.048 & 0.092 & 0.056 \\
                     &                     & 3    & 0.058  & 0.056  & 0.046  & 0.086  & 0.056 & 10 & 0.058 & 0.06  & 0.044 & 0.102 & 0.062 \\ \hline
\multirow{9}{*}{$\mathcal{H}_0^{3}$}   & \multirow{3}{*}{3}  & 0.5  & 0.05   & 0.05   & 0.038  & 0.124  & 0.652 & 3  & 0.046 & 0.046 & 0.034 & 0.102 & 0.586 \\
                     &                     & 1    & 0.048  & 0.05   & 0.032  & 0.128  & 0.546 & 5  & 0.04  & 0.052 & 0.024 & 0.11  & 0.598 \\
                     &                     & 3    & 0.058  & 0.048  & 0.046  & 0.102  & 0.534 & 10 & 0.036 & 0.036 & 0.028 & 0.13  & 0.704 \\
                     & \multirow{3}{*}{20} & 0.5  & 0.048  & 0.056  & 0.036  & 0.124  & 0.652 & 3  & 0.044 & 0.048 & 0.036 & 0.102 & 0.424 \\
                     &                     & 1    & 0.038  & 0.04   & 0.018  & 0.102  & 0.56  & 5  & 0.05  & 0.05  & 0.042 & 0.102 & 0.476 \\
                     &                     & 3    & 0.062  & 0.062  & 0.036  & 0.134  & 0.556 & 10 & 0.04  & 0.052 & 0.03  & 0.116 & 0.528 \\
                     & \multirow{3}{*}{50} & 0.5  & 0.074  & 0.08   & 0.06   & 0.096  & 0.218 & 3  & 0.06  & 0.048 & 0.046 & 0.114 & 0.31  \\
                     &                     & 1    & 0.038  & 0.05   & 0.032  & 0.114  & 0.236 & 5  & 0.094 & 0.076 & 0.064 & 0.114 & 0.31  \\
                     &                     & 3    & 0.054  & 0.054  & 0.034  & 0.102  & 0.276 & 10 & 0.044 & 0.04  & 0.026 & 0.098 & 0.31  \\ \hline
\multirow{9}{*}{$\mathcal{H}_0^{4}$}   & \multirow{3}{*}{3}  & 0.5  & 0.034  & 0.034  & 0.024  & 0.092  & 0.192 & 3  & 0.046 & 0.042 & 0.03  & 0.07  & 0.356 \\
                     &                     & 1    & 0.048  & 0.04   & 0.03   & 0.084  & 0.28  & 5  & 0.032 & 0.044 & 0.014 & 0.08  & 0.416 \\
                     &                     & 3    & 0.052  & 0.05   & 0.038  & 0.066  & 0.448 & 10 & 0.044 & 0.056 & 0.032 & 0.074 & 0.44  \\
                     & \multirow{3}{*}{20} & 0.5  & 0.04   & 0.032  & 0.034  & 0.1    & 0.158 & 3  & 0.054 & 0.06  & 0.034 & 0.082 & 0.252 \\
                     &                     & 1    & 0.04   & 0.034  & 0.028  & 0.098  & 0.276 & 5  & 0.06  & 0.046 & 0.044 & 0.088 & 0.294 \\
                     &                     & 3    & 0.056  & 0.044  & 0.04   & 0.06   & 0.444 & 10 & 0.06  & 0.05  & 0.038 & 0.054 & 0.29  \\
                     & \multirow{3}{*}{50} & 0.5  & 0.026  & 0.022  & 0.018  & 0.098  & 0.07  & 3  & 0.03  & 0.034 & 0.018 & 0.05  & 0.152 \\
                     &                     & 1    & 0.046  & 0.048  & 0.026  & 0.106  & 0.096 & 5  & 0.048 & 0.062 & 0.036 & 0.098 & 0.174 \\
                     &                     & 3    & 0.066  & 0.058  & 0.05   & 0.096  & 0.21  & 10 & 0.054 & 0.046 & 0.038 & 0.092 & 0.198 \\ \hline
\multirow{9}{*}{$\mathcal{H}_0^{5}$}  & \multirow{3}{*}{3}  & 0.5  & 0.052  & 0.05   & 0.034  & 0.096  & 0.164 & 3  & 0.048 & 0.056 & 0.034 & 0.132 & 0.312 \\
                     &                     & 1    & 0.044  & 0.042  & 0.032  & 0.118  & 0.21  & 5  & 0.046 & 0.048 & 0.032 & 0.118 & 0.322 \\
                     &                     & 3    & 0.06   & 0.056  & 0.046  & 0.13   & 0.474 & 10 & 0.044 & 0.044 & 0.026 & 0.106 & 0.376 \\
                     & \multirow{3}{*}{20} & 0.5  & 0.048  & 0.032  & 0.04   & 0.112  & 0.148 & 3  & 0.048 & 0.028 & 0.032 & 0.11  & 0.2   \\
                     &                     & 1    & 0.052  & 0.04   & 0.032  & 0.104  & 0.224 & 5  & 0.036 & 0.05  & 0.026 & 0.078 & 0.18  \\
                     &                     & 3    & 0.064  & 0.068  & 0.046  & 0.124  & 0.502 & 10 & 0.062 & 0.062 & 0.04  & 0.122 & 0.242 \\
                     & \multirow{3}{*}{50} & 0.5  & 0.052  & 0.058  & 0.04   & 0.112  & 0.056 & 3  & 0.058 & 0.062 & 0.044 & 0.08  & 0.118 \\
                     &                     & 1    & 0.052  & 0.05   & 0.042  & 0.112  & 0.104 & 5  & 0.078 & 0.068 & 0.05  & 0.082 & 0.16  \\
                     &                     & 3    & 0.056  & 0.056  & 0.036  & 0.116  & 0.268 & 10 & 0.05  & 0.034 & 0.026 & 0.096 & 0.178 \\ \hline
\end{tabular}
\end{table}



\subsection{Dynamic settings}\label{subsec:sequentialsimulation}
In this subsection, we first construct two simulation environments, both sharing a common time horizon $T=24$ and state dimension $d=3$: one featuring a linear data generating process (DGP) and the other a nonlinear DGP. These environments allow us to systematically examine the performance of different ATE test statistics. In both environments, we implement a switchback design: $A_{i,t}=1-A_{i,t-1}$ for all $t>1$ and $A_{i,1}=1-A_{i-1,T}$ for all $i>1$, with $A_{1,1} \sim \text{Uniform}\{0,1\}$.

\textbf{Linear DGP:} Data is generated based on model 
\begin{equation}\label{eqn:linear_MDP}
	\begin{split}
		& Y_{i,t}=\alpha_t + \beta_t^\top X_{i,t}+\gamma_{t}A_{i,t} +e_{i,t} ,\\
		& X_{i,t+1}=\phi_t +\Phi_t X_{i,t} +\Gamma_t A_{i,t} +E_{i,t}. 
	\end{split}
\end{equation}
As outlined in \citet{luo2022policy}, the ATE can be expressed as
\begin{eqnarray}\label{ate_est_formula}
	\textrm{ATE}=\frac{1}{T}\sum_{t=1}^T \gamma_t+\frac{1}{T}\sum_{t=2}^T \beta_t^\top \Big[ \sum_{k=1}^{t-1} (\Phi_{t-1}\Phi_{t-2}\ldots\Phi_{k+1}) \Gamma_k \Big],
\end{eqnarray}
where the product $\Phi_{t-1}\ldots \Phi_{k+1}$ is treated as an identity matrix if $t-1<k+1$.

The number of days $n$ used in our simulations varies, selected from the set {$\{100, 150, 300\}$} and time horizons is $T=24$. The initial state for each day is drawn from a 3-dimensional multivariate normal distribution with zero mean and an identity covariance matrix. The coefficients for these models are specified as : $ \{\Phi_t^{(j_1,j_2)}\}_{t,j_1,j_2}\stackrel{i.i.d.}{\sim }  U[-0.3, 0.3]$, $ \{\Gamma_t^{(j)}\}_{t, j} \stackrel{i.i.d.}{\sim }  N(0, 0.5 \delta)$ and 
\begin{equation*}
	\begin{split}
		&\{\alpha_t\}_t \stackrel{i.i.d.}{\sim }  \begin{cases}
			U[-1,-0.5]&\text{ with probability 0.5 } \\
			U[0.5, 1]&\text{ with probability 0.5 }
		\end{cases} ,	\{\beta_t^{(j)}\}_{t,j}  \stackrel{i.i.d.}{\sim }  \begin{cases}
			U[-0.3, -0.1] & \text{ with probability 0.5 } \\
			U[0.1, 0.3] & \text{ with probability 0.5 } 
		\end{cases}, \\
		&\{\phi_t^{(j)}\}_{t,j} \stackrel{i.i.d.}{\sim }  \begin{cases}
			U[-1,-0.5] & \text{ with probability 0.5 } \\
			U[0.5, 1]  & \text{ with probability 0.5 }
		\end{cases},\quad     \{\gamma_t\}_t \stackrel{i.i.d.}{\sim }  \begin{cases}
	0& \text{ if $\delta$ =0}  \\
		U[0.1\delta , 0.1+0.8 \delta]  & \text{ else }
		\end{cases}.
	\end{split}
\end{equation*}
Here, the superscript $j$ denotes the $j$th component of each vector, while $(j_1,j_2)$ indicates the element in the $j_1$th row and $j_2$th column of each matrix. Note that the experimental hyperparameter $\delta$ represents the strength of the treatment policy: a larger value of $\delta$ leads to larger treatment effects, and when $\delta = 0$, the ATE equals zero. It is selected from $\{0, 0.015, 0.055, 0.1, 0.15, 0.25\}$. And the actions are generated according to a switchback design, where the time span for each switch is set to $1$.

Both the reward error $e_t$ and the residual in the state regression model $E_t = X_{t+1} - \Mean (X_{t+1}|A_t,X_t)$
are set to mean zero Gaussian noises. Specifically, $e_t=\eta_t + \varepsilon_t$ where $\{\varepsilon_t\}_t$ are i.i.d. Gaussian errors $N(0, 1.5)$, and $\{\eta_t\}_t$ are random effects with an autoregressive covariance function: $\sigma_{\eta}(t_1,t_2) = 1.5\rho^{|t_1-t_2|}$. The parameter $\rho=0.5$. The sequence $\{E_t\}_t$ is set to an i.i.d. multivariate Gaussian error process, with a covariance matrix 1.5 times the identity matrix, and it is independent of $\{e_t\}_t$.

\textbf{NonLinear DGP:} We consider the nonlinear reward function: $ r_t(a,x)=\alpha_t +2\beta_t^\top [\sin (x)+\cos(x)]^2 +3(\beta_t^\top x) \gamma_t a +[ a \gamma_t+\cos(a \gamma_t) ]^2$, where the sine, cosine, and square functions are applied element-wise to each component of the vector. The state regression function remains linear and identical to the one presented in \eqref{eqn:linear_MDP}. All model parameters, including $\{\alpha_t\}_t$, $\{\beta_t\}_t$, $\{\gamma_t\}_t$, $\{\Gamma_t\}_t$, $\{\phi_t\}_t$, $n$ and $T$, are the same as those in the setting of Linear DGP, with the exception of $\Phi_t^{(j_1, j_2)}\stackrel{i.i.d.}{\sim }  U[-0.6, 0.6]$ for $j_1, j_2=1, 2, 3$. 

\textbf{Inference:} For each setting, we conduct $1000$ repetitions to evaluate the empirical type-I error rates and
powers of the following three tests: (i) P-TAB; (ii) TAB; (iii) DRL. All tests are performed at a significance level of 0.05 and employ
2-fold cross-fitting to construct the pseudo outcomes and to form the test statistics. 

\begin{figure}[t]
	\centering
	\begin{minipage}{0.8\linewidth}
		\centering
		\includegraphics[width=\linewidth]{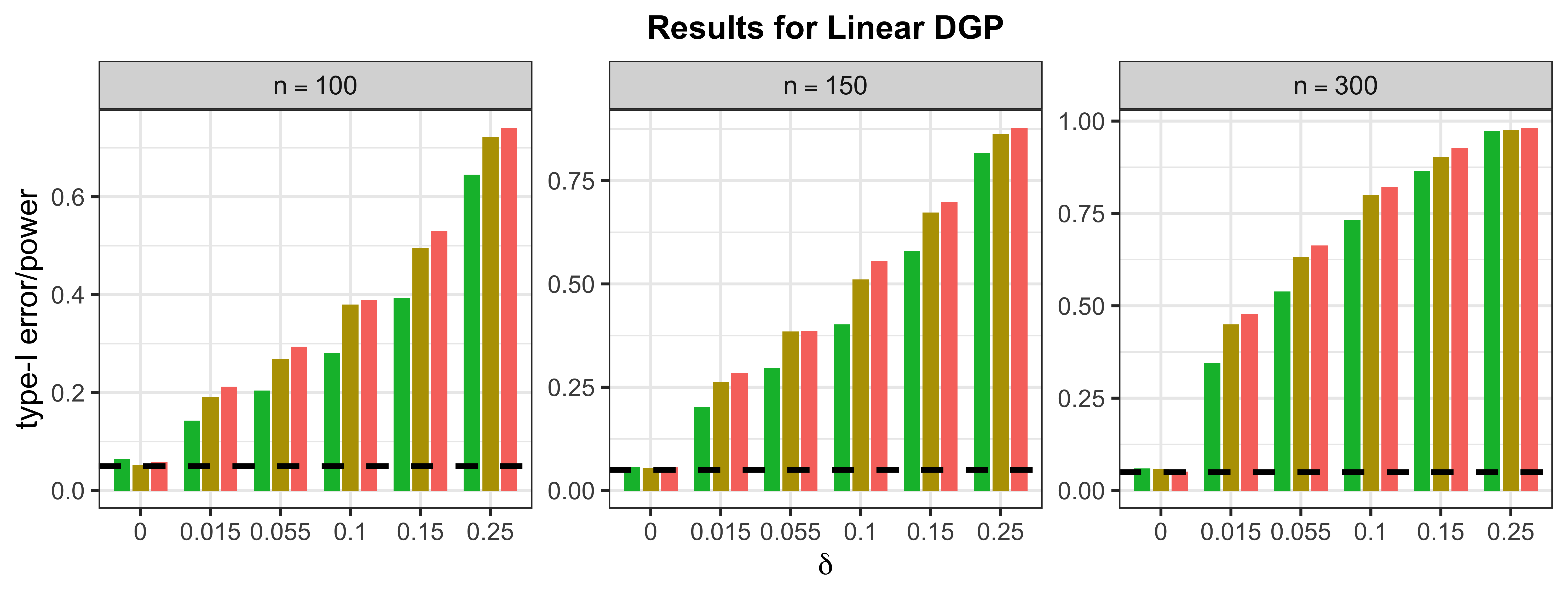}
	\end{minipage}
	\hspace{0.1cm} 
	\begin{minipage}{0.8\linewidth}
		\centering
		\includegraphics[width=\linewidth]{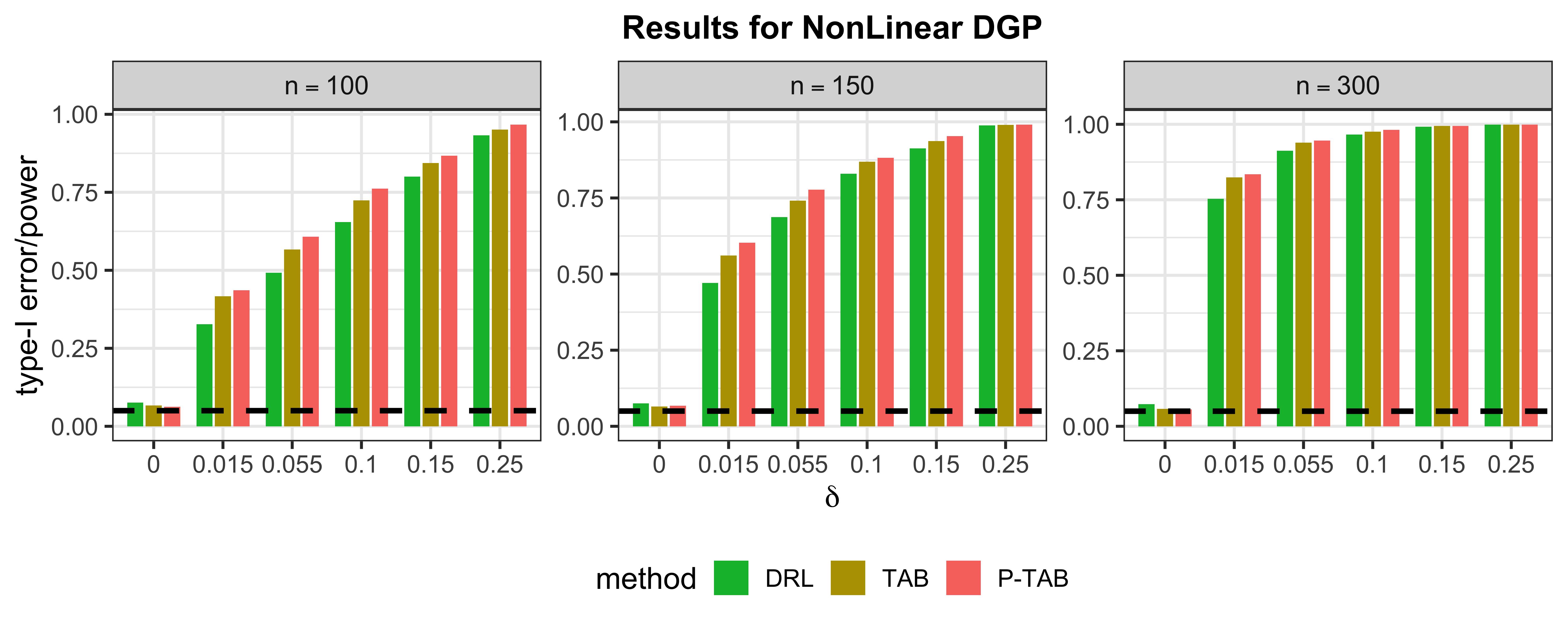}
	\end{minipage}
	
	\caption{\small Type-I errors and powers of different methods under the Linear/NonLinear DGPs.}
	\label{fig:DGP_all_probs}
\end{figure}

\textbf{Results: } As illustrated in Figure~\ref{fig:DGP_all_probs} and Table \ref{tab:DGP}, when $\delta=0$, the proposed test statistics P-TAB and TAB have smaller sizes (type-I errors) compared to the DRL method. As $\delta$ increases, the power of these statistics follows the order: $\text{P-TAB} > \text{TAB} > \text{DRL}$, indicating that our proposed statistics achieve greater power than the standard DRL method. Moreover, as either $\delta$ or the sample size $n$ increases, the power of all test statistics improves. At $\delta=0$, all test statistics maintain a nominal size level, approximately $0.05$. These results highlight the advantages of our proposed methods.

\begin{table}[ht]
\caption{\small Type-I error rates and statistical powers for all methods, with relative improvements over the DRL method defined as:
		$\text{P-TAB\_improv} = \frac{\text{Power}_\text{P-TAB} - \text{Power}_\text{DRL}}{\text{Power}_\text{DRL}}$ and 
		$\text{TAB\_improv} = \frac{\text{Power}_\text{TAB} - \text{Power}_\text{DRL}}{\text{Power}_\text{DRL}}$, respectively.
	 } \label{tab:DGP}
\begin{tabular}{cccccccc}
\hline
DGP                         & $n$                    & $\delta$                & P-TAB                & TAB                  & DR                   & P-TAB\_improv        & TAB\_improv          \\ \hline
\multirow{18}{*}{Linear}    & \multirow{6}{*}{100} & 0                    & 0.058                & 0.052                & 0.065                & -               & -              \\
                            &                      & 0.015                & 0.212                & 0.191                & 0.143                & 0.483                & 0.336                \\
                            &                      & 0.055                & 0.294                & 0.269                & 0.204                & 0.441                & 0.319                \\
                            &                      & 0.1                  & 0.389                & 0.38                 & 0.281                & 0.384                & 0.352                \\
                            &                      & 0.15                 & 0.53                 & 0.495                & 0.394                & 0.345                & 0.256                \\
                            &                      & 0.25                 & 0.741                & 0.722                & 0.645                & 0.149                & 0.119                \\
                            & \multirow{6}{*}{150} & 0                    & 0.056                & 0.054                & 0.057                & -               & -               \\
                            &                      & 0.015                & 0.284                & 0.263                & 0.203                & 0.399                & 0.296                \\
                            &                      & 0.055                & 0.387                & 0.385                & 0.297                & 0.303                & 0.296                \\
                            &                      & 0.1                  & 0.556                & 0.511                & 0.402                & 0.383                & 0.271                \\
                            &                      & 0.15                 & 0.699                & 0.673                & 0.58                 & 0.205                & 0.160                \\
                            &                      & 0.25                 & 0.878                & 0.862                & 0.817                & 0.075                & 0.055                \\
                            & \multirow{6}{*}{300} & 0                    & 0.051                & 0.059                & 0.06                 & -               & -               \\
                            &                      & 0.015                & 0.477                & 0.45                 & 0.345                & 0.383                & 0.304                \\
                            &                      & 0.055                & 0.663                & 0.632                & 0.539                & 0.230                & 0.173                \\
                            &                      & 0.1                  & 0.821                & 0.8                  & 0.732                & 0.122                & 0.093                \\
                            &                      & 0.15                 & 0.927                & 0.903                & 0.864                & 0.073                & 0.045                \\
                            &                      & 0.25                 & 0.982                & 0.975                & 0.973                & 0.009                & 0.002                \\ \hline
\multirow{18}{*}{NonLinear} & \multirow{6}{*}{100} & 0                    & 0.063                & 0.067                & 0.076                & -             & -               \\
                            &                      & 0.015                & 0.436                & 0.417                & 0.327                & 0.333                & 0.275                \\
                            &                      & 0.055                & 0.608                & 0.567                & 0.492                & 0.236                & 0.152                \\
                            &                      & 0.1                  & 0.762                & 0.724                & 0.655                & 0.163                & 0.105                \\
                            &                      & 0.15                 & 0.867                & 0.844                & 0.8                  & 0.084                & 0.055                \\
                            &                      & 0.25                 & 0.967                & 0.951                & 0.933                & 0.036                & 0.019                \\
                            & \multirow{6}{*}{150} & 0                    & 0.068                & 0.065                & 0.075                & -               & -               \\
                            &                      & 0.015                & 0.603                & 0.561                & 0.471                & 0.280                & 0.191                \\
                            &                      & 0.055                & 0.777                & 0.741                & 0.687                & 0.131                & 0.079                \\
                            &                      & 0.1                  & 0.882                & 0.869                & 0.83                 & 0.063                & 0.047                \\
                            &                      & 0.15                 & 0.953                & 0.937                & 0.913                & 0.044                & 0.026                \\
                            &                      & 0.25                 & 0.991                & 0.99                 & 0.988                & 0.003                & 0.002                \\
                            & \multirow{6}{*}{300} & 0                    & 0.056                & 0.058                & 0.073                & -               & -               \\
                            &                      & 0.015                & 0.835                & 0.824                & 0.753                & 0.109                & 0.094                \\
                            &                      & 0.055                & 0.946                & 0.939                & 0.912                & 0.037                & 0.030                \\
                            &                      & 0.1                  & 0.982                & 0.976                & 0.966                & 0.017                & 0.010                \\
                            &                      & 0.15                 & 0.995                & 0.995                & 0.992                & 0.003                & 0.003                \\
                            &                      & 0.25                 & 0.999                & 0.999                & 0.999                & 0.000                & 0.000                \\ \hline

\end{tabular}
\end{table}

\section{Evaluation of order dispatch policies} \label{sec:suppOrderDispatch}

This section details the wild bootstrap procedures used to construct the simulation environments in Section \ref{subsec: real_data_based_exps} and provides the associated numerical results.

\subsection{Bootstrap-based simulation}\label{algrithombootstrap}

\begin{algorithm}[h] \label{alg:bootstrap}
	\caption{Bootstrap-based simulation.}\label{algo:res_bootstrap}
	\SetAlgoLined 
	\KwIn{Real data $\left\lbrace (X_{i,t}, Y_{i,t}): 1 \leq i \leq N; 1 \leq t \leq T \right\rbrace$, 
    policy improvement $\lambda$, time intervals per day $T$, bootstrapped sample size $n$, and simulation repetitions $B$.}
	\KwOut{$P$-values for the simulated datasets, type-I error rates/powers.}
	\textbf{Initialization:} Calculating the least square estimates $\{
	\widehat{\alpha} \}_t $,  $\{
	\widehat{\beta}_t \}_t $,  $\{
	\widehat{\phi}_t \}_t $,  $\{
	\widehat{\Phi}_t \}_t $  in the model \eqref{eqn:linear_MDP}, calibrating the treatment effect parameters  $\{ \widehat{\gamma}_t \}_t $ and  $\{
	\widehat{\Gamma}_t \}_t $ by the given improvement $\lambda$, and computing the residuals for the reward model and state regression model by model  \eqref{eqn:residuals}\;
	\For{$b=1$ \KwTo $B$}{
		
		\textbf{1.} Generate $n\times T$ treatment assignments by adopting a switchback design in which the assigned treatment alternates at each time step, i.e., $A_{i,t}=1-A_{i,t-1}$ for all $t>1$ and $A_{i,1}=1-A_{i-1,T}$ for all $i>1$, with the initial action $A_{1,1}$ being generated uniformly at random.
        
		\textbf{2.} Simulate state-reward trajectories $\{(\widehat{X}_{i,t},\widehat{Y}_{i,t})\}_{t}$ for each $i\leq n$ using model \eqref{eqn:linear_MDPsimulated}.
        
		\textbf{3.}   Apply the Algorithm \ref{alg:p-tab-DR} of the main text to calculate the p-value for the simulated dataset.
	}
    Report type-I errors/powers as the proportion of null hypothesis rejections across $B$ simulation replications.
\end{algorithm}

Following the bootstrap-based simulation procedure in \citet{li2024combining} and \citet{wenunraveling2025}, we generate simulated datasets from observational data. For each simulated dataset, variables $X_t$ and $Y_t$ are generated according to model (\ref{eqn:linear_MDP}), while $A_t$ is assigned via the switchback design described in the main text. 

Specifically, for each original dataset, the parameters $\{\alpha_t\}_t$, $\{\beta_t\}_t$, $\{\phi_t\}_t$, and $\{\Phi_t\}_t$ in model (\ref{eqn:linear_MDP}) require estimation since they are unobserved. Using observational data, we estimate these regression coefficients via ridge regression, where the regularization parameter is selected by minimizing the generalized cross-validation criterion \citep{wahba1975smoothing}. This produces estimators $\{\widehat{\alpha}_t\}_t$, $\{\widehat{\beta}_t\}_t$, $\{\widehat{\phi}_t\}_t$, and $\{\widehat{\Phi}_t\}_t$. However, the parameters $\{\gamma_t\}_t$ and $\{\Gamma_t\}_t$ remain unidentifiable because $A_t = 0$ holds almost surely. We determine them using pre-specified policy improvement levels quantified by $\lambda$. As detailed in the main text, we examine six distinct $\lambda$ values. For each $\lambda$, we calibrate $\{\gamma_t\}_t$ and $\{\Gamma_t\}_t$ to achieve the target policy improvement through Equation (\ref{ate_est_formula}), assuming equal direct and indirect effects.

Leveraging the estimated parameters $\{\widehat{\alpha}_t\}_t$, $\{\widehat{\beta}_t\}_t$, $\{\widehat{\phi}_t\}_t$, $\{\widehat{\Phi}_t\}_t$ and the specified parameters $\{\widehat{\gamma}_t\}_t$ and $\{\widehat{\Gamma}_t\}_t$, we sequentially generate simulated $X_t$ and $Y_t$, denoted by $\widehat{Y}_t$ and $\widehat{X}_t$, via
\begin{equation}\label{eqn:linear_MDPsimulated}
	\begin{split}
		& \widehat Y_{i,t}=\widehat\alpha_t + \widehat\beta_t^\top \widehat X_{i,t}+\widehat\gamma_{t}A_{i,t} +\xi_i \widehat e_{i,t} ,\\
		& \widehat X_{i,t+1}=\widehat\phi_t +\widehat\Phi_t \widehat X_{i,t} +\widehat\Gamma_t A_{i,t} +\xi_i \widehat E_{i,t},
	\end{split}
\end{equation}
for $i = 1,2,\dots,n$ and $t=1,2,\dots,T$, where each initial state $\widehat{X}_{i,1}$ is bootstrapped from the 40 initial states in the original dataset with replacement, $\xi_i$ is independently sampled from the standard Gaussian distribution, $\widehat e_{i,t} = \widehat e_{r,t}$ and $\widehat E_{i,t} = \widehat E_{r,t}$ are the residuals in the reward and state regression models: 
\begin{equation}\label{eqn:residuals}
	\widehat{e}_{r,t}=Y_{r,t}-\widehat{\alpha}_t-X_{r,t}^\top \widehat{\beta}_t, \quad \widehat{E}_{r,t}=X_{r,t+1}-\widehat{\phi}_{t}- \widehat{\Phi}_t X_{r,t},
\end{equation}
with $r$ being the index of the bootstrapped initial state $\widehat X_{i,1}$ in the original dataset, 
and $n \in \{30,50,100\}$ is the sample size for simulated dataset. This simulation method preserves the error covariance structure of the original dataset in the simulated data.

For each simulated dataset, we compute the corresponding $p$-value using Algorithm \ref{alg:p-tab-DR} in the main text. 
We then estimate the test's empirical rejection rate as the proportion of null hypothesis rejections across all $B$ simulation repetitions, calculated separately for each combination of policy improvement $\lambda$ and sample size $n$. The complete methodology is formalized in Algorithm \ref{alg:bootstrap}.

\subsection{Numerical results}

\begin{table}[t]
\caption{Type-I error rates and statistical powers for all methods, with relative improvements over the DRL method defined as:
		$\text{P-TAB\_improv} = \frac{\text{Power}_\text{P-TAB} - \text{Power}_\text{DRL}}{\text{Power}_\text{DRL}}$ and 
		$\text{TAB\_improv} = \frac{\text{Power}_\text{TAB} - \text{Power}_\text{DRL}}{\text{Power}_\text{DRL}}$, respectively.} \label{tab:Real_data}
\begin{tabular}{cccccccc}
\hline
Dataset            & $n$                    & $\lambda$            & P-TAB                & TAB                  & DR                   & P-TAB\_improv        & TAB\_improv          \\ \hline
\multirow{18}{*}{First dataset}  & \multirow{6}{*}{30}  & 0                    & 0.015                & 0.012                & 0.016                & -               & -               \\
                             &                      & 0.002                & 0.046                & 0.045                & 0.032                & 0.438                & 0.406                \\
                             &                      & 0.004                & 0.109                & 0.106                & 0.083                & 0.313                & 0.277                \\
                             &                      & 0.01                 & 0.284                & 0.283                & 0.241                & 0.178                & 0.174                \\
                             &                      & 0.02                 & 0.581                & 0.574                & 0.514                & 0.130                & 0.117                \\
                             &                      & 0.05                 & 0.908                & 0.894                & 0.874                & 0.039                & 0.023                \\
                             & \multirow{6}{*}{50}  & 0                    & 0.008                & 0.009                & 0.014                & -               & -               \\
                             &                      & 0.002                & 0.084                & 0.076                & 0.057                & 0.474                & 0.333                \\
                             &                      & 0.004                & 0.195                & 0.196                & 0.15                 & 0.300                & 0.307                \\
                             &                      & 0.01                 & 0.537                & 0.513                & 0.455                & 0.180                & 0.127                \\
                             &                      & 0.02                 & 0.864                & 0.858                & 0.818                & 0.056                & 0.049                \\
                             &                      & 0.05                 & 0.992                & 0.992                & 0.989                & 0.003                & 0.003                \\
                             & \multirow{6}{*}{100} & 0                    & 0.021                & 0.027                & 0.02                 & -                & -                \\
                             &                      & 0.002                & 0.203                & 0.192                & 0.146                & 0.390                & 0.315                \\
                             &                      & 0.004                & 0.457                & 0.45                 & 0.378                & 0.209                & 0.190                \\
                             &                      & 0.01                 & 0.866                & 0.861                & 0.82                 & 0.056                & 0.050                \\
                             &                      & 0.02                 & 0.996                & 0.995                & 0.991                & 0.005                & 0.004                \\
                             &                      & 0.05                 & 1                    & 1                    & 1                    & 0.000                & 0.000                \\ \hline
\multirow{18}{*}{Second dataset} & \multirow{6}{*}{30}  & 0                    & 0.016                & 0.02                 & 0.024                & -              & -               \\
                             &                      & 0.002                & 0.035                & 0.039                & 0.028                & 0.250                & 0.393                \\
                             &                      & 0.004                & 0.059                & 0.059                & 0.041                & 0.439                & 0.439                \\
                             &                      & 0.01                 & 0.206                & 0.196                & 0.16                 & 0.288                & 0.225                \\
                             &                      & 0.02                 & 0.41                 & 0.396                & 0.364                & 0.126                & 0.088                \\
                             &                      & 0.05                 & 0.825                & 0.812                & 0.789                & 0.046                & 0.029                \\
                             & \multirow{6}{*}{50}  & 0                    & 0.018                & 0.015                & 0.022                & -               & -               \\
                             &                      & 0.002                & 0.047                & 0.048                & 0.036                & 0.306                & 0.333                \\
                             &                      & 0.004                & 0.09                 & 0.084                & 0.066                & 0.364                & 0.273                \\
                             &                      & 0.01                 & 0.327                & 0.313                & 0.272                & 0.202                & 0.151                \\
                             &                      & 0.02                 & 0.619                & 0.599                & 0.54                 & 0.146                & 0.109                \\
                             &                      & 0.05                 & 0.953                & 0.944                & 0.93                 & 0.025                & 0.015                \\
                             & \multirow{6}{*}{100} & 0                    & 0.013                & 0.013                & 0.016                & -               & -               \\
                             &                      & 0.002                & 0.075                & 0.072                & 0.049                & 0.531                & 0.469                \\
                             &                      & 0.004                & 0.163                & 0.148                & 0.126                & 0.294                & 0.175                \\
                             &                      & 0.01                 & 0.56                 & 0.54                 & 0.455                & 0.231                & 0.187                \\
                             &                      & 0.02                 & 0.892                & 0.866                & 0.821                & 0.086                & 0.055                \\
                             &                      & 0.05                 & 1                    & 1                    & 0.999                & 0.001                & 0.001                \\ \hline

\end{tabular}
\end{table}

Table \ref{tab:Real_data} presents type I error rates and power estimates. For $\lambda >0$, the two rightmost columns quantify power improvements of P-TAB and TAB over DRL, respectively.

\end{document}